\documentclass[10pt,twocolumn,letterpaper]{article}

\usepackage{arxiv}
\usepackage{times}
\usepackage{epsfig}
\usepackage{graphicx}
\usepackage{amsmath}
\usepackage{amssymb}

\usepackage{algorithm}
\usepackage{algpseudocode}
\algrenewcommand\algorithmicindent{.9em}%

\usepackage{comment}
\usepackage[british,UKenglish,USenglish,english,american]{babel}
\usepackage{multirow}
\usepackage{xcolor}
\usepackage{mathtools}

\usepackage{booktabs}
\usepackage{makecell}
\usepackage{subfigure}
\usepackage[pagebackref=true,breaklinks=true,letterpaper=true,colorlinks,bookmarks=false]{hyperref}

\begin{document}

\title{Deformable ConvNets v2: More Deformable, Better Results}

\author{Xizhou Zhu$^{1,2}$\thanks{This work is done when Xizhou Zhu is an intern at Microsoft Research Asia.}\quad Han Hu$^2$ \quad Stephen Lin$^2$ \quad Jifeng Dai$^2$ \vspace{8pt}\\
    $^1$University of Science and Technology of China\\
        $^2$Microsoft Research Asia\\
        {\tt\small ezra0408@mail.ustc.edu.cn} \\
        {\tt\small \{hanhu,stevelin,jifdai\}@microsoft.com} \\
}
\maketitle

\begin{abstract}

The superior performance of Deformable Convolutional Networks arises from its ability to adapt to the geometric variations of objects. Through an examination of its adaptive behavior, we observe that while the spatial support for its neural features conforms more closely than regular ConvNets to object structure, this support may nevertheless extend well beyond the region of interest, causing features to be influenced by irrelevant image content. To address this problem, we present a reformulation of Deformable ConvNets that improves its ability to focus on pertinent image regions, through increased modeling power and stronger training.
The modeling power is enhanced through a more comprehensive integration of deformable convolution within the network, and by introducing a modulation mechanism that expands the scope of deformation modeling. To effectively harness this enriched modeling capability, we guide network training via a proposed feature mimicking scheme that helps the network to learn features that reflect the object focus and classification power of R-CNN features. With the proposed contributions, this new version of Deformable ConvNets yields significant performance gains over the original model and produces leading results on the COCO benchmark for object detection and instance segmentation. 

\end{abstract}

\section{Introduction}

Geometric variations due to scale, pose, viewpoint and part deformation present a major challenge in object recognition and detection. The current state-of-the-art method for addressing this issue is {\it Deformable Convolutional Networks} (DCNv1)~\cite{dai2017deformable}, which introduces two modules that aid CNNs in modeling such variations. One of these modules is {\it deformable convolution}, in which the grid sampling locations of standard convolution are each offset by displacements learned with respect to the preceding feature maps. The other is {\it deformable RoIpooling}, where offsets are learned for the bin positions in RoIpooling~\cite{girshick2015fast}. The incorporation of these modules into a neural network gives it the ability to adapt its feature representation to the configuration of an object, specifically by deforming its sampling and pooling patterns to fit the object's structure. With this approach, large improvements in object detection accuracy are obtained.

Towards understanding Deformable ConvNets, the authors visualized the induced changes in receptive field, via the arrangement of offset sampling positions in PASCAL VOC images~\cite{everingham2010pascal}. It is found that samples for an activation unit tend to cluster around the object on which it lies. However, the coverage over an object is inexact, exhibiting a spread of samples beyond the area of interest. In a deeper analysis of spatial support using images from the more challenging COCO dataset~\cite{lin2014microsoft}, we observe that such behavior becomes more pronounced. These findings suggest that greater potential exists for learning deformable convolutions.

In this paper, we present a new version of Deformable ConvNets, called {\it Deformable ConvNets v2} (DCNv2), with enhanced modeling power for learning deformable convolutions. This increase in modeling capability comes in two complementary forms. The first is the expanded use of deformable convolution layers within the network. Equipping more convolutional layers with offset learning capacity allows DCNv2 to control sampling over a broader range of feature levels. The second is a modulation mechanism in the deformable convolution modules, where each sample not only undergoes a learned offset, but is also modulated by a learned feature amplitude. The network module is thus given the ability to vary both the spatial distribution and the relative influence of its samples.

To fully exploit the increased modeling capacity of DCNv2, effective training is needed. Inspired by work on knowledge distillation in neural networks~\cite{ba2014deep,hinton2015distilling}, we make use of a teacher network for this purpose, where the teacher provides guidance during training. We specifically utilize R-CNN~\cite{girshick2014rich} as the teacher. Since it is a network trained for classification on cropped image content, R-CNN learns features unaffected by irrelevant information outside the region of interest. To emulate this property, DCNv2 incorporates a feature mimicking loss into its training, which favors learning of features consistent to those of R-CNN. In this way, DCNv2 is given a strong training signal for its enhanced deformable sampling.

With the proposed changes, the deformable modules remain lightweight and can easily be incorporated into existing network architectures. Specifically, we incorporate DCNv2 into the Faster R-CNN~\cite{ren2015faster} and Mask R-CNN~\cite{he2017mask} systems, with a variety of backbone networks. Extensive experiments on the COCO benchmark demonstrate the significant improvement of DCNv2 over DCNv1 for object detection and instance segmentation. The code for DCNv2 will be released.

\section{Analysis of Deformable ConvNet Behavior}

\subsection{Spatial Support Visualization}

To better understand the behavior of Deformable ConvNets, we visualize the spatial support of network nodes by their effective receptive fields~\cite{luo2017understanding}, effective sampling locations, and error-bounded saliency regions. These three modalities provide different and complementary perspectives on the underlying image regions that contribute to a node's response.

\vspace{0.5em}
\noindent\textbf{Effective receptive fields}
Not all pixels within the receptive field of a network node contribute equally to its response. The differences in these contributions are represented by an {\em effective receptive field}, whose values are calculated as the gradient of the node response with respect to intensity perturbations of each image pixel~\cite{luo2017understanding}. We utilize the effective receptive field to examine the relative influence of individual pixels on a network node, but note that this measure does not reflect the structured influence of full image regions.

\vspace{0.5em}
\noindent\textbf{Effective sampling / bin locations}
In~\cite{dai2017deformable}, the sampling locations of (stacked) convolutional layers and the sampling bins in RoIpooling layers are visualized for understanding the behavior of Deformable ConvNets. However, the relative contributions of these sampling locations to the network node are not revealed. We instead visualize {\em effective sampling locations} that incorporate this information, computed as the gradient of the network node with respect to the sampling / bin locations, so as to understand their contribution strength.

\vspace{0.5em}
\noindent\textbf{Error-bounded saliency regions}
The response of a network node will not change if we remove image regions that do not influence it, as demonstrated in recent research on image saliency~\cite{zhou2016learning,zintgraf2017visualizing,fong2017interpretable,dabkowski2017real}. Based on this property, we can determine a node's support region as the smallest image region giving the same response as the full image, within a small error bound. We refer to this as the {\it error-bounded saliency region}, which can be found by progressively masking parts of the image and computing the resulting node response, as described in more detail in the Appendix. The error-bounded saliency region facilitates comparison of support regions from different networks.

\subsection{Spatial Support of Deformable ConvNets}

\begin{figure}[t] 
  \centering 
  \subfigure[regular conv]{ 
    \includegraphics[width=1.025\linewidth]{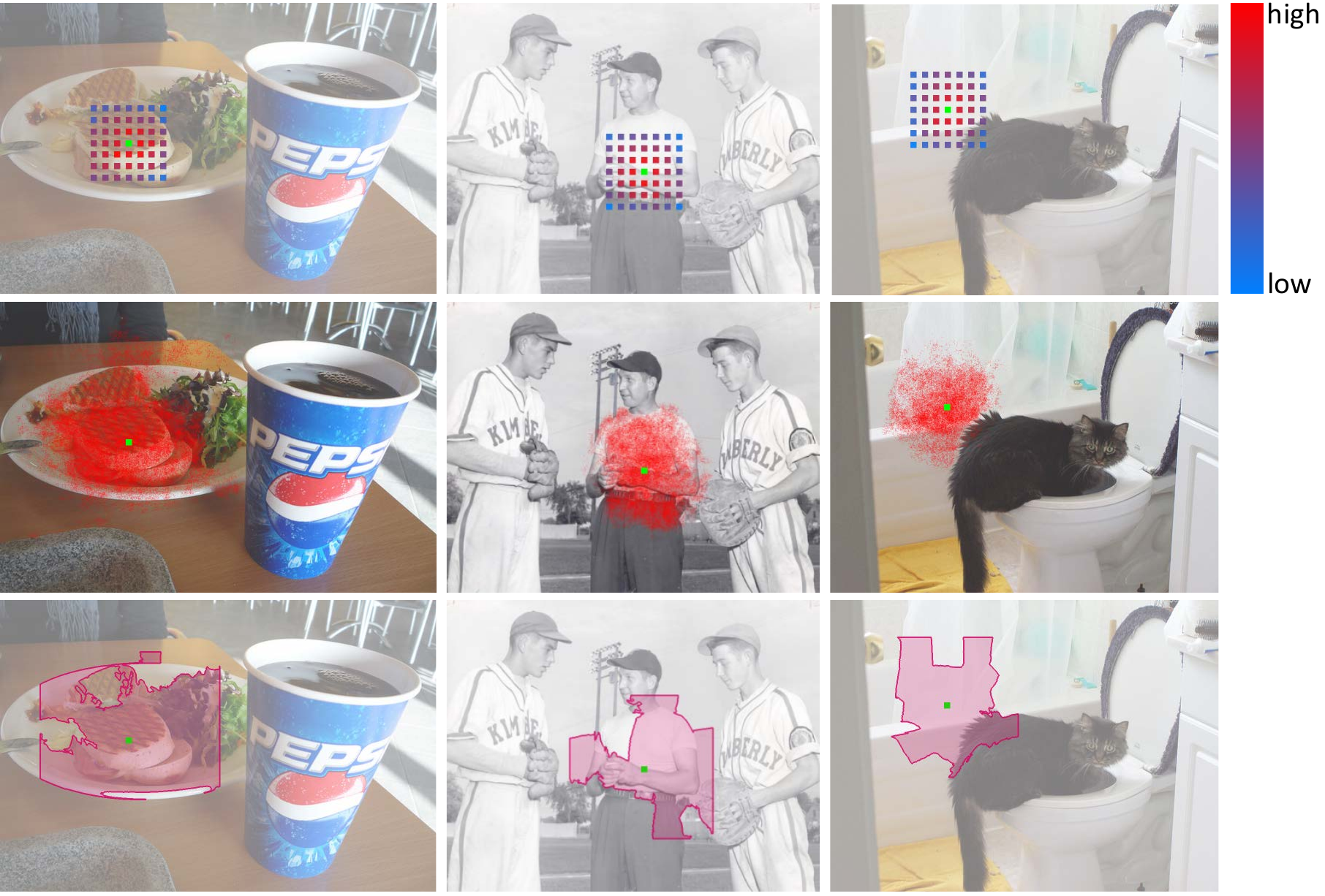}} 
  \hspace{1in} 
  \subfigure[deformable conv@conv5 stage (DCNv1)]{ 
    \includegraphics[width=1.025\linewidth]{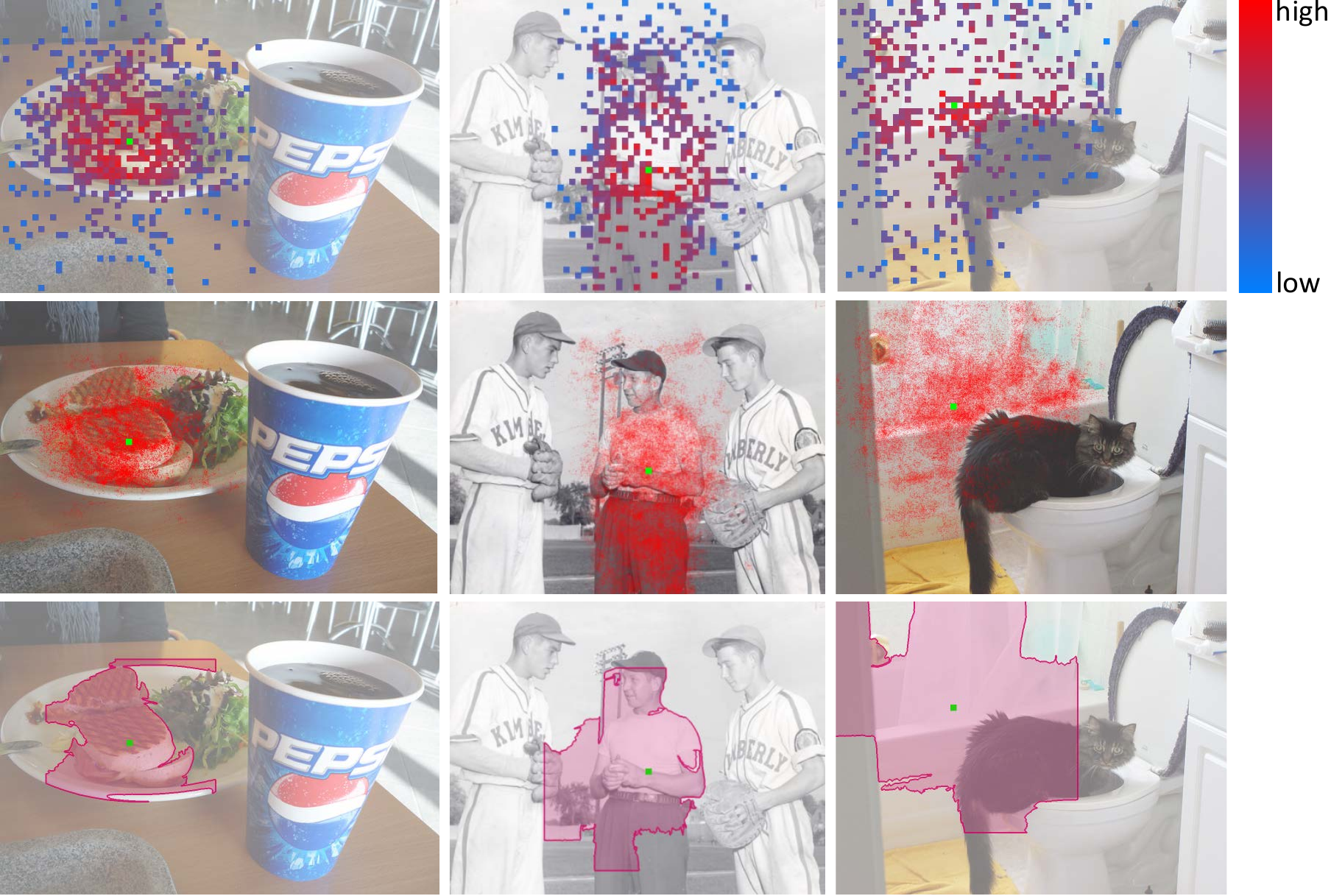}}
  \subfigure[modulated deformable conv@conv3$\sim$5 stages (DCNv2)]{ 
    \includegraphics[width=1.025\linewidth]{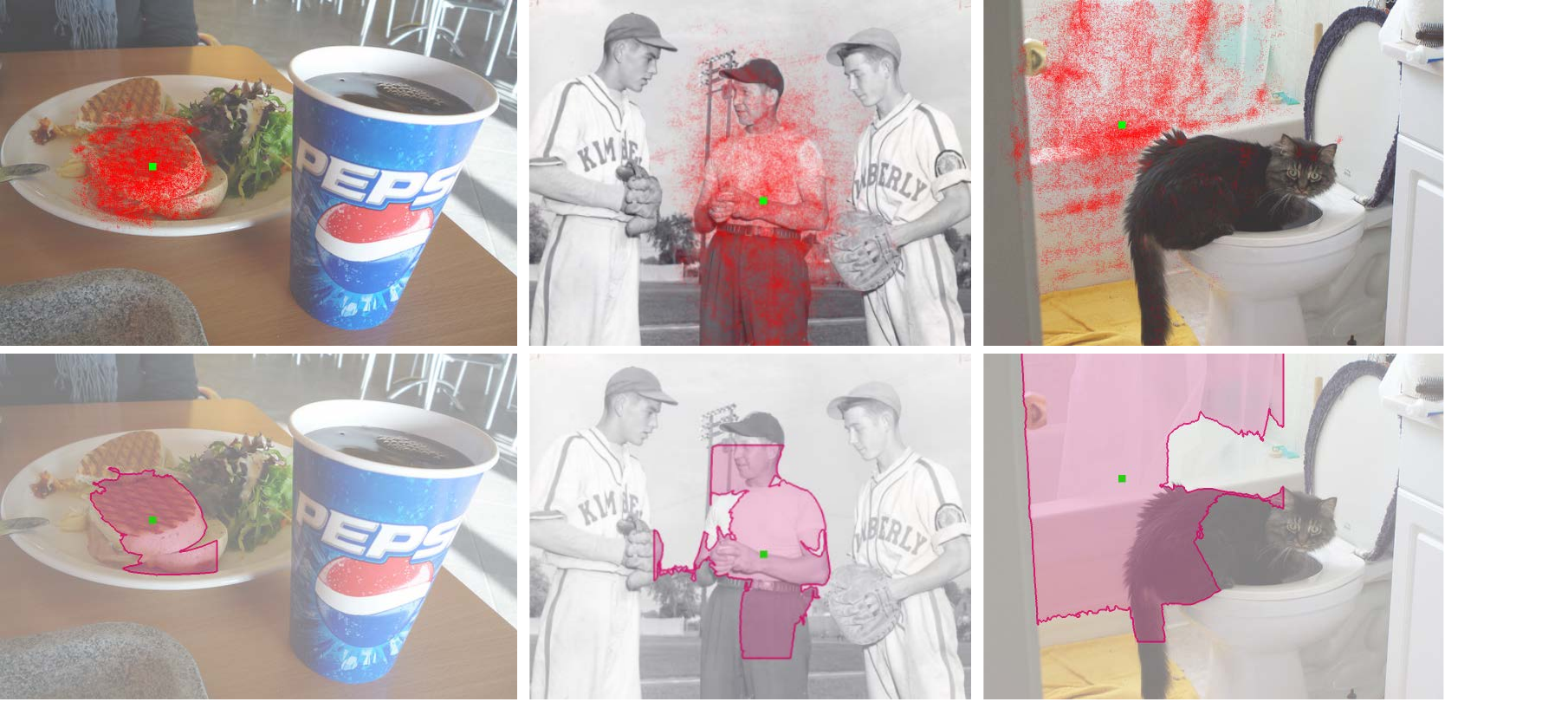}}
  \caption{Spatial support of nodes in the last layer of the conv5 stage in a regular ConvNet, DCNv1 and DCNv2. The regular ConvNet baseline is Faster R-CNN + ResNet-50. In each subfigure, the effective sampling locations, effective receptive field,  and error-bounded saliency regions are shown from the top to the bottom rows. Effective sampling locations are omitted in (c) as they are similar to those in (b), providing limited additional information. The visualized nodes (green points) are on a small object (left), a large object (middle), and the background (right).}
  \label{fig:visualize_regular_deformable_conv}
  \vspace{-1em}
\end{figure}

We analyze the visual support regions of Deformable ConvNets in object detection. The regular ConvNet we employ as a baseline consists of a Faster R-CNN + ResNet-50~\cite{he2016deep} object detector with aligned RoIpooling\footnote{Aligned RoIpooling is called RoIAlign in~\cite{he2017mask}. We use the term ``aligned RoIpooling" in this paper to more clearly describe it in the context of other related terms.}~\cite{he2017mask}. All the convolutional layers in ResNet-50 are applied on the whole input image. The effective stride in the conv5 stage is reduced from 32 to 16 pixels to increase feature map resolution. The RPN~\cite{ren2015faster} head is added on top of the conv4 features of ResNet-101. On top of the conv5 features we add the Fast R-CNN head~\cite{girshick2015fast}, which is composed of aligned RoIpooling and two fully-connected ({\it fc}) layers, followed by the classification and bounding box regression branches. We follow the procedure in~\cite{dai2017deformable} to turn the object detector into its deformable counterpart. The three layers of $3\times 3$ convolutions in the conv5 stage are replaced by deformable convolution layers. Also, the aligned RoIpooling layer is replaced by deformable RoIPooling. Both networks are trained and visualized on the COCO benchmark. It is worth mentioning that when the offset learning rate is set to zero, the Deformable Faster R-CNN detector degenerates to regular Faster R-CNN with aligned RoIpooling.

Using the three visualization modalities, we examine the spatial support of nodes in the last layer of the conv5 stage in Figure~\ref{fig:visualize_regular_deformable_conv} (a)$\sim$(b). The sampling locations analyzed in~\cite{dai2017deformable} are also shown. From these visualizations, we make the following observations:

1. Regular ConvNets can model geometric variations to some extent, as evidenced by the changes in spatial support with respect to image content. Thanks to the strong representation power of deep ConvNets, the network weights are learned to accommodate some degree of geometric transformation.

2. By introducing deformable convolution, the network's ability to model geometric transformation is considerably enhanced, even on the challenging COCO benchmark. The spatial support adapts much more to image content, with nodes on the foreground having support that covers the whole object, while nodes on the background have expanded support that encompasses greater context. However, the range of spatial support may be inexact, with the effective receptive field and error-bounded saliency region of a foreground node including background areas irrelevant for detection. 

3. The three presented types of spatial support visualizations are more informative than the sampling locations used in~\cite{dai2017deformable}. This can be seen, for example, with regular ConvNets, which have fixed sampling locations along a grid, but actually adapt its effective spatial support via network weights. The same is true for Deformable ConvNets, whose predictions are jointly affected by learned offsets and network weights. Examining sampling locations alone, as done in \cite{dai2017deformable}, can result in misleading conclusions about Deformable ConvNets. 

\begin{figure*}[t] 
  \centering 
  \subfigure[aligned RoIpooling, with deformable conv@conv5 stage]{ 
    \includegraphics[width=0.48\linewidth]{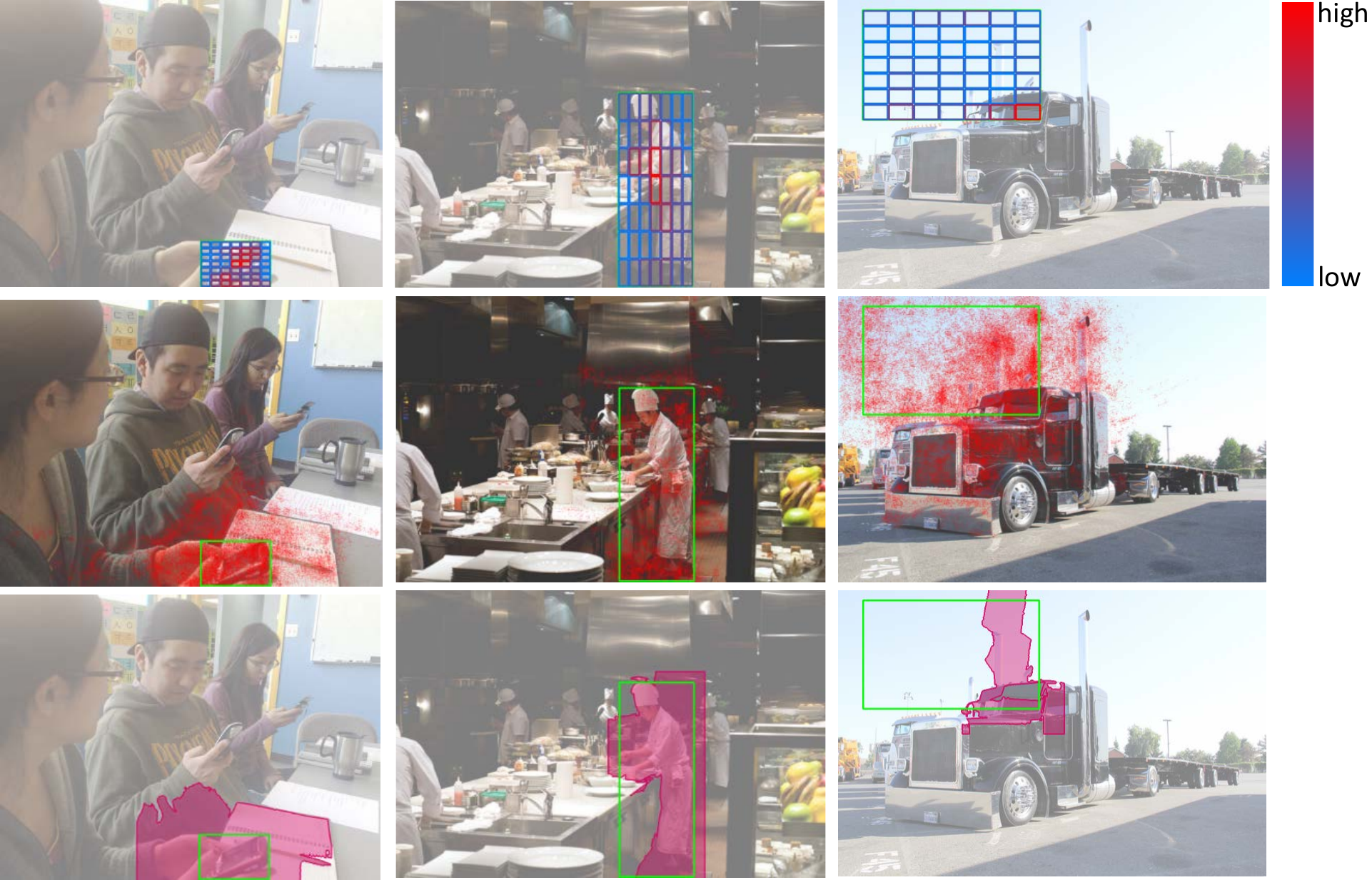}}  
  \subfigure[deformable RoIpooling, with deformable conv@conv5 stage (DCNv1)]{
    \includegraphics[width=0.48\linewidth]{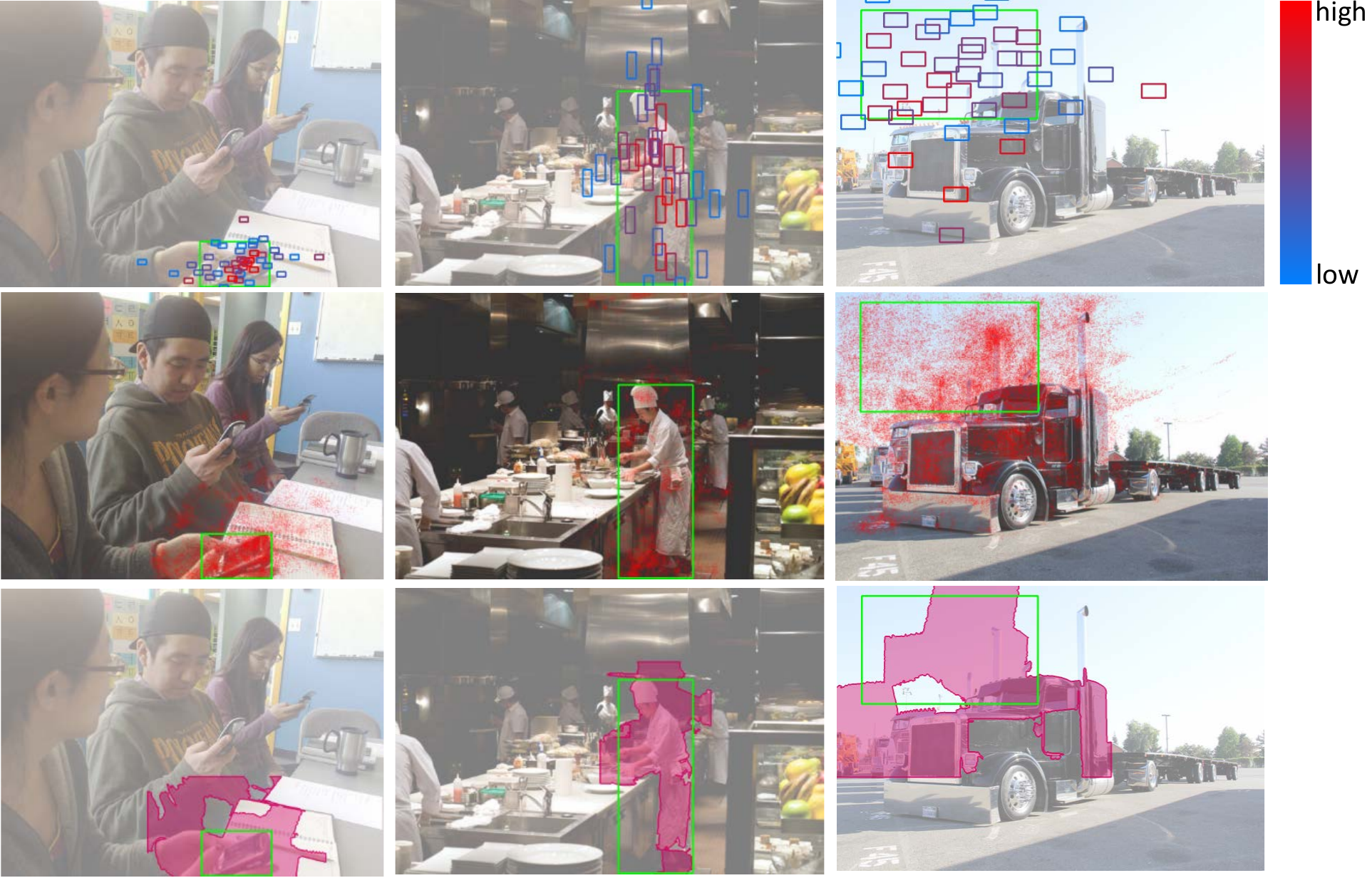}}
  \subfigure[modulated deformable RoIpooling, with modulated deformable conv@conv3$\sim$5 stages]{
    \includegraphics[width=0.48\linewidth]{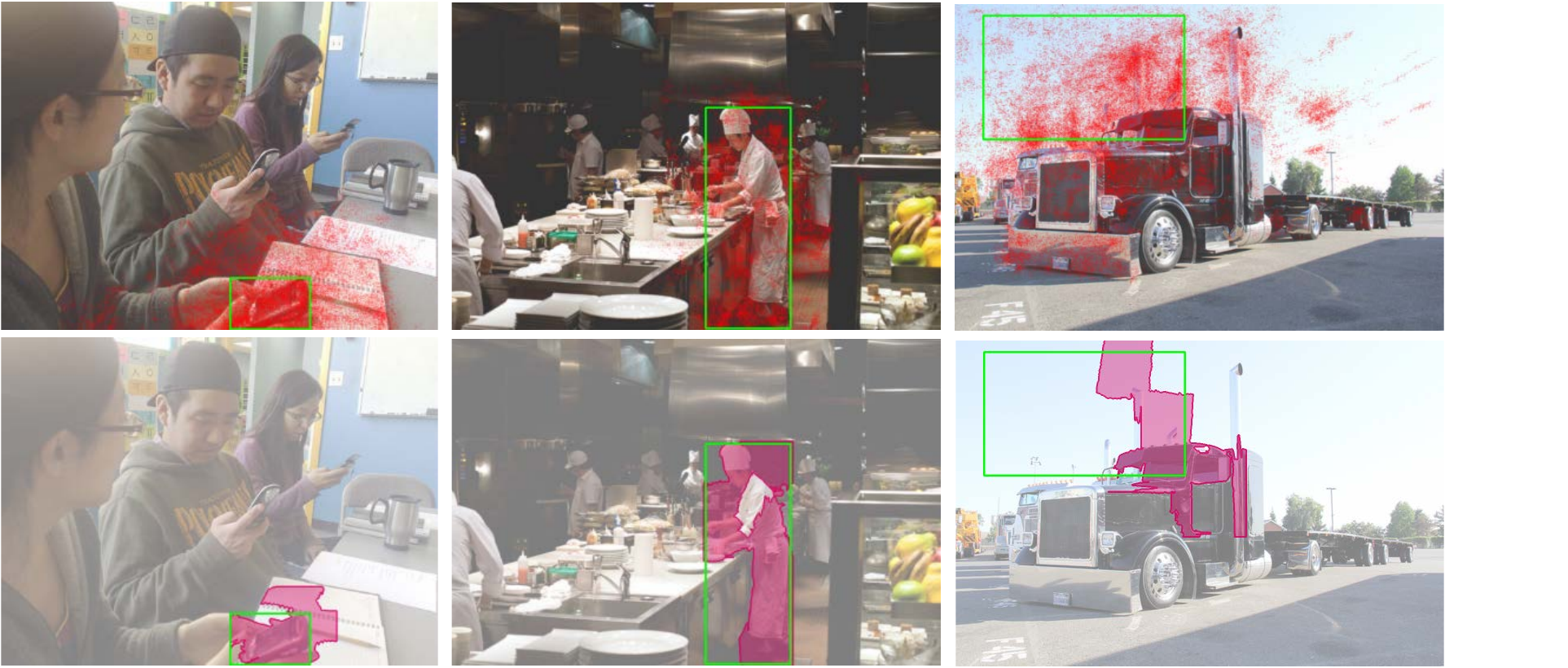}}
  \subfigure[with R-CNN feature mimicking on setting (c) (DCNv2)]{
    \includegraphics[width=0.48\linewidth]{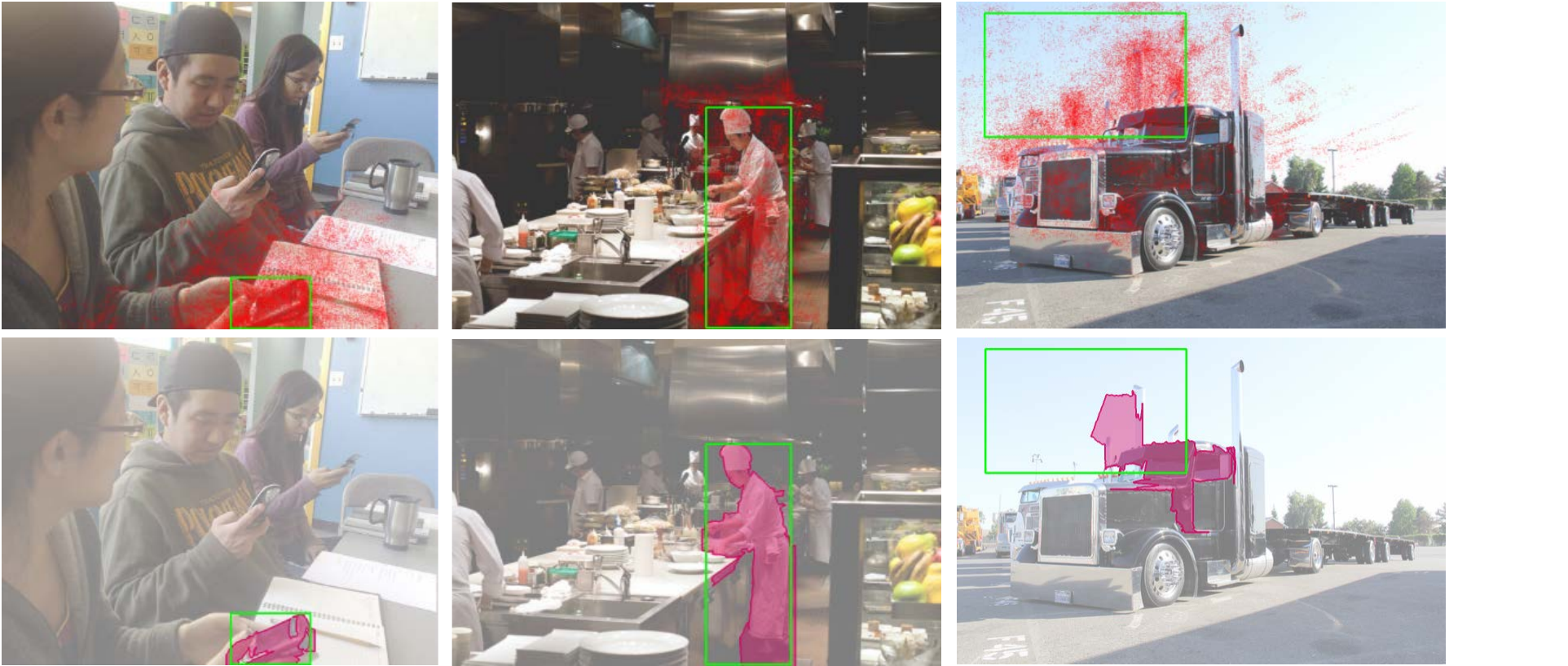}}
  \begin{minipage}[c]{0.5\textwidth}
    \subfigure[with R-CNN feature mimicking in regular ConvNet]{
      \includegraphics[width=0.96\linewidth]{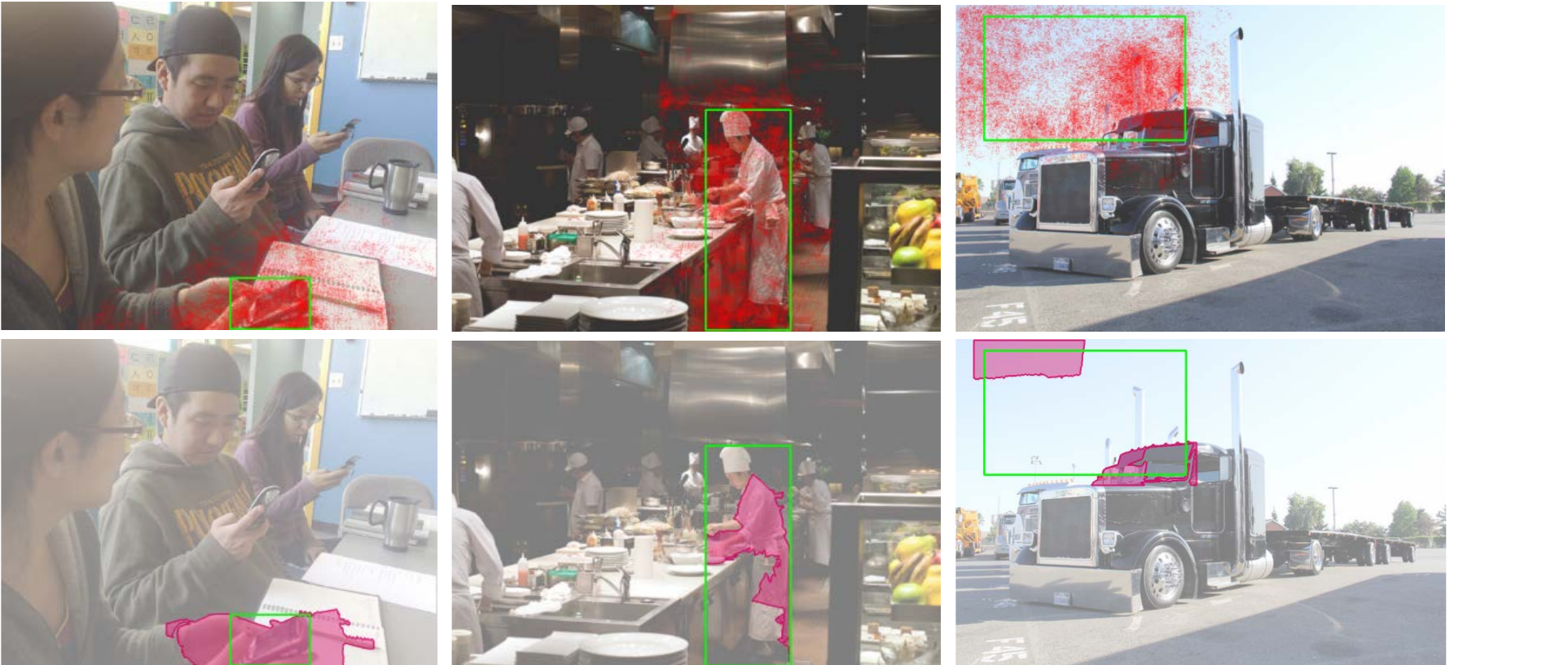}}
  \end{minipage}\hfill
  \begin{minipage}[c]{0.5\textwidth}
    \caption{Spatial support of the 2{\it fc} node in the per-RoI detection head, directly followed by the classification and the bounding box regression branches. Visualization is conducted on a regular ConvNet, DCNv1 and DCNv2. The regular ConvNet baseline is Faster R-CNN + ResNet-50. In each subfigure, the effective bin locations, effective receptive fields, and error-bounded saliency regions are shown from the top to the bottom rows, except for (c)$\sim$(e) where the effective bin locations are omitted as they provide little additional understanding over those in (a)$\sim$(b). The input RoIs (green boxes) are on a small object (left), a large object (middle), and the background (right).}
    \label{fig:visualize_regular_deformable_roi}
  \end{minipage}
  \vspace{-1em}
\end{figure*}

Figure~\ref{fig:visualize_regular_deformable_roi} (a)$\sim$(b) display the spatial support of the 2{\it fc} node in the per-RoI detection head, which is directly followed by the classification and the bounding box regression branches. The visualization of effective bin locations suggests that bins on the object foreground generally receive larger gradients from the classification branch, and thus exert greater influence on prediction. This observation holds for both aligned RoIpooling and Deformable RoIpooling. In Deformable RoIpooling, a much larger proportion of bins cover the object foreground than in aligned RoIpooling, thanks to the introduction of learnable bin offsets. Thus, more information from relevant bins is available for the downstream Fast R-CNN head. Meanwhile, the error-bounded saliency regions in both aligned RoIpooling and Deformable RoIpooling are not fully focused on the object foreground, which suggests that image content outside of the RoI affects the prediction result. According to a recent study~\cite{cheng2018revisiting}, such feature interference could be harmful for detection.

While it is evident that Deformable ConvNets have markedly improved ability to adapt to geometric variation in comparison to regular ConvNets, it can also be seen that their spatial support may extend beyond the region of interest. We thus seek to upgrade Deformable ConvNets so that they can better focus on pertinent image content and deliver greater detection accuracy.

\section{More Deformable ConvNets}

To improve the network's ability to adapt to geometric variations, we present changes to boost its modeling power and to help it take advantage of this increased capability.

\subsection{Stacking More Deformable Conv Layers}

Encouraged by the observation that Deformable ConvNets can effectively model geometric transformation on challenging benchmarks, we boldly replace more regular conv layers by their deformable counterparts. We expect that by stacking more deformable conv layers, the geometric transformation modeling capability of the entire network can be further strengthened.

In this paper, deformable convolutions are applied in all the $3\times 3$ conv layers in stages conv3, conv4, and conv5 in ResNet-50. Thus, there are 12 layers of deformable convolution in the network. In contrast, just three layers of deformable convolution are used in~\cite{dai2017deformable}, all in the conv5 stage. It is observed in~\cite{dai2017deformable} that performance saturates when stacking more than three layers for the relatively simple and small-scale PASCAL VOC benchmark. Also, misleading offset visualizations on COCO may have hindered further exploration on more challenging benchmarks. In experiments, we observe that utilizing deformable layers in the conv3-conv5 stages achieves the best tradeoff between accuracy and efficiency for object detection on COCO. See Section~\ref{sec:exp_enriched} for details.

\subsection{Modulated Deformable Modules}

To further strengthen the capability of Deformable ConvNets in manipulating spatial support regions, a modulation mechanism is introduced. With it, the Deformable ConvNets modules can not only adjust offsets in perceiving input features, but also modulate the input feature amplitudes from different spatial locations / bins. In the extreme case, a module can decide not to perceive signals from a particular location / bin by setting its feature amplitude to zero. Consequently, image content from the corresponding spatial location will have considerably reduced or no impact on the module output. Thus, the modulation mechanism provides the network module another dimension of freedom to adjust its spatial support regions. 

Given a convolutional kernel of $K$ sampling locations, let $w_k$ and $p_k$ denote the weight and pre-specified offset for the $k$-th location, respectively. For example, $K=9$ and $p_k\in \{(-1,-1), (-1, 0), \ldots, (1, 1)\}$ defines a $3\times 3$ convolutional kernel of dilation 1. Let $x(p)$ and $y(p)$ denote the features at location $p$ from the input feature maps $x$ and output feature maps $y$, respectively. The modulated deformable convolution can then be expressed as
\begin{equation}
y(p) = \sum_{k=1}^{K} w_k \cdot x(p+p_k+\Delta p_k)\cdot \Delta m_k,
\end{equation}
where $\Delta p_k$ and $\Delta m_k$ are the learnable offset and modulation scalar for the $k$-th location, respectively. The modulation scalar $\Delta m_k$ lies in the range $[0, 1]$, while $\Delta p_k$ is a real number with unconstrained range. As $p+p_k+\Delta p_k$ is fractional, bilinear interpolation is applied as in~\cite{dai2017deformable} in computing $x(p+p_k+\Delta p_k)$. Both $\Delta p_k$ and $\Delta m_k$ are obtained via a separate convolution layer applied over the same input feature maps $x$. This convolutional layer is of the same spatial resolution and dilation as the current convolutional layer. The output is of $3K$ channels, where the first $2K$ channels correspond to the learned offsets $\{\Delta p_k\}_{k=1}^{K}$, and the remaining $K$ channels are further fed to a sigmoid layer to obtain the modulation scalars $\{\Delta m_k\}_{k=1}^{K}$. The kernel weights in this separate convolution layer are initialized to zero. Thus, the initial values of $\Delta p_k$ and $\Delta m_k$ are 0 and 0.5, respectively.
 The learning rates of the added conv layers for offset and modulation learning are set to 0.1 times those of the existing layers. 

The design of modulated deformable RoIpooling is similar. Given an input RoI, RoIpooling divides it into $K$ spatial bins (\eg $7\times 7$). Within each bin, sampling grids of even spatial intervals are applied (\eg $2\times 2$). The sampled values on the grids are averaged to compute the bin output. Let $\Delta p_k$ and $\Delta m_k$ be the learnable offset and modulation scalar for the $k$-th bin. The output binning feature $y(k)$ is computed as
\begin{equation}
y(k) = \sum_{j=1}^{n_k}x(p_{kj}+\Delta p_k)\cdot \Delta m_k/n_k,
\end{equation}
where $p_{kj}$ is the sampling location for the $j$-th grid cell in the $k$-th bin, and $n_k$ denotes the number of sampled grid cells. Bilinear interpolation is applied to obtain features $x(p_{kj}+\Delta p_k)$.
The values of $\Delta p_k$ and $\Delta m_k$ are produced by a sibling branch on the input feature maps. In this branch, RoIpooling generates features on the RoI, followed by two {\it fc} layers of 1024-D (initialized with Gaussian distribution of standard derivation of 0.01). On top of that, an additional {\it fc} layer produces output of $3K$ channels (weights initialized to be zero). The first $2K$ channels are the normalized learnable offsets, where element-wise multiplications with the RoI's width and height are computed to obtain $\{\Delta p_k\}_{k=1}^{K}$. The remaining $K$ channels are normalized by a sigmoid layer
to produce $\{\Delta m_k\}_{k=1}^{K}$. The learning rates of the added {\it fc} layers for offset learning are the same as those of the existing layers. 

\subsection{R-CNN Feature Mimicking}
\label{sec:mimic}
As observed in Figure~\ref{fig:visualize_regular_deformable_roi}, the error-bounded saliency region of a per-RoI classification node can stretch beyond the RoI for both regular ConvNets and Deformable ConvNets. Image content outside of the RoI may thus affect the extracted features and consequently degrade the final results of object detection. 

In~\cite{cheng2018revisiting}, the authors find redundant context to be a plausible source of detection error for Faster R-CNN. Together with other motivations (\eg, to share fewer features between the classification and bounding box regression branches), the authors propose to combine the classification scores of Faster R-CNN and R-CNN to obtain the final detection score. Since R-CNN classification scores are focused on cropped image content from the input RoI, incorporating them would help to alleviate the redundant context problem and improve detection accuracy. However, the combined system is slow because both the Faster-RCNN and R-CNN branches need to be applied in both training and inference.

Meanwhile, Deformable ConvNets are powerful in adjusting spatial support regions. For Deformable ConvNets v2 in particular, the modulated deformable RoIpooling module could simply set the modulation scalars of bins in a way that excludes redundant context. However, our experiments in Section~\ref{sec:exp_mimicking} show that even with modulated deformable modules, such representations cannot be learned well through the standard Faster R-CNN training procedure.
We suspect that this is because the conventional Faster R-CNN training loss cannot effectively drive the learning of such representations. Additional guidance is needed to steer the training.

Motivated by recent work on feature mimicking~\cite{ba2014deep,hinton2015distilling,li2017mimicking}, we incorporate a feature mimic loss on the per-RoI features of Deformable Faster R-CNN to force them to be similar to R-CNN features extracted from cropped images. This auxiliary training objective is intended to drive Deformable Faster R-CNN to learn more ``focused'' feature representations like R-CNN. We note that, based on the visualized spatial support regions in Figure~\ref{fig:visualize_regular_deformable_roi}, a focused feature representation may well not be optimal for negative RoIs on the image background. For background areas, more context information may need to be considered so as not to produce false positive detections. Thus, the feature mimic loss is enforced only on positive RoIs that sufficiently overlap with ground-truth objects.

\begin{figure}[t]
\centering
\includegraphics[width=0.45\textwidth]{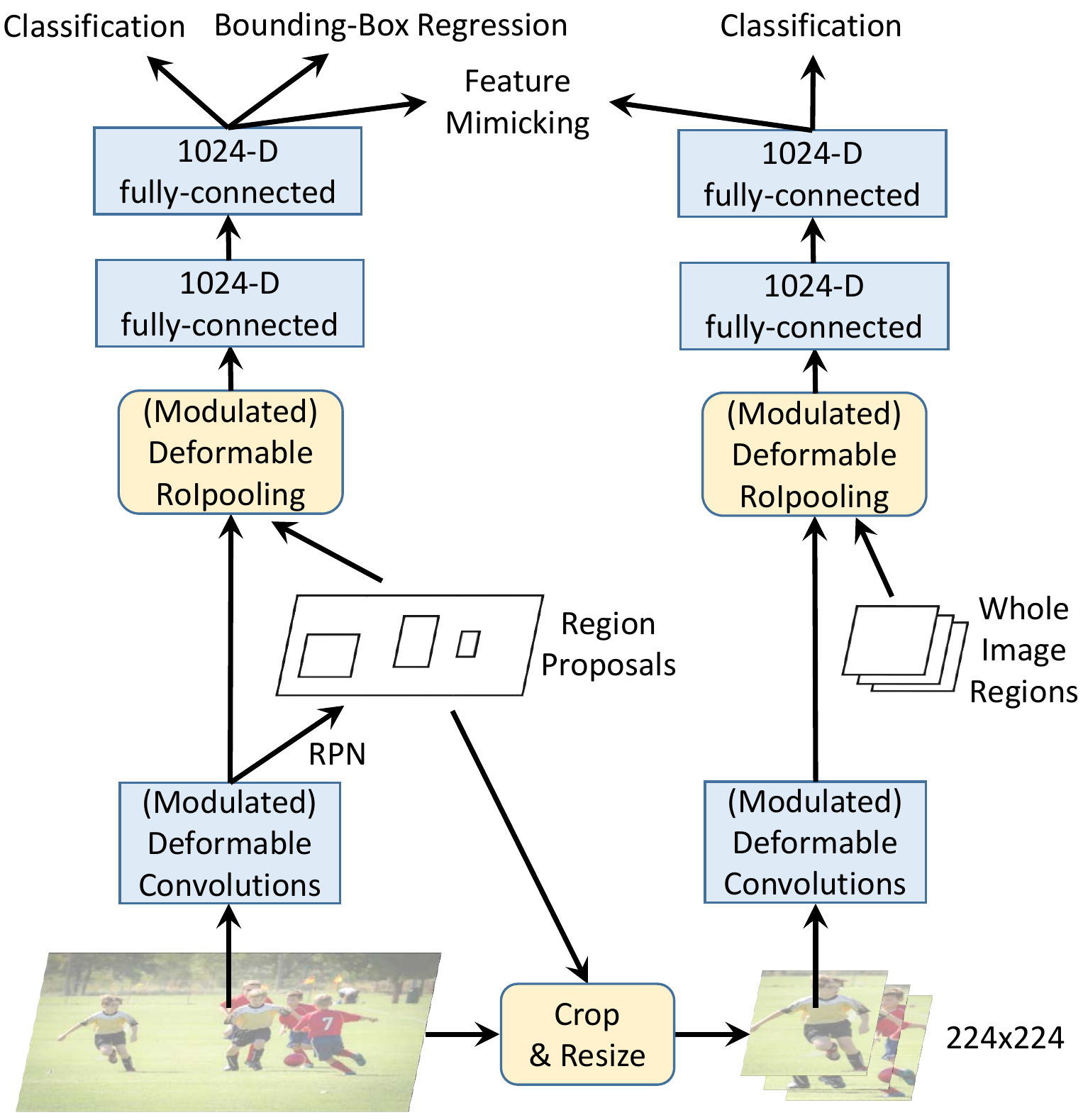}
\caption{Network training with R-CNN feature mimicking.}
\label{fig:mimic_rcnn_training}
\vspace{-1em}
\end{figure}

The network architecture for training Deformable Faster R-CNN is presented in Figure~\ref{fig:mimic_rcnn_training}. In addition to the Faster R-CNN network, an additional R-CNN branch is added for feature mimicking. Given an RoI $b$ for feature mimicking, the image patch corresponding to it is cropped and resized to $224 \times 224$ pixels. In the R-CNN branch, the backbone network operates on the resized image patch and produces feature maps of $14 \times 14$ spatial resolution.  A (modulated) deformable RoIpooling layer is applied on top of the feature maps, where the input RoI covers the whole resized image patch (top-left corner at $(0, 0)$, and height and width are 224 pixels). After that, 2 {\it fc} layers of 1024-D are applied, producing an R-CNN feature representation for the input image patch, denoted by $f_{\text{RCNN}}(b)$. A $(C+1)$-way Softmax classifier follows for classification, where $C$ denotes the number of foreground categories, plus one for background. The feature mimic loss is enforced between the R-CNN feature representation $f_{\text{RCNN}}(b)$ and the counterpart in Faster R-CNN, $f_{\text{FRCNN}}(b)$, which is also 1024-D and is produced by the 2 {\it fc} layers in the Fast R-CNN head. The feature mimic loss is defined on the cosine similarity between $f_{\text{RCNN}}(b)$ and $f_{\text{FRCNN}}(b)$, computed as
\begin{equation}
L_{\text{mimic}} = \sum_{b \in \Omega} [1-\cos(f_{\text{RCNN}}(b), f_{\text{FRCNN}}(b))],
\end{equation}
where $\Omega$ denotes the set of RoIs sampled for feature mimic training. In the SGD training, given an input image, 32 positive region proposals generated by RPN are randomly sampled into $\Omega$. A cross-entropy classification loss is enforced on the R-CNN classification head, also computed on the RoIs in $\Omega$. Network training is driven by the feature mimic loss and the R-CNN classification loss, together with the original loss terms in Faster R-CNN. The loss weights of the two newly introduced loss terms are 0.1 times those of the original Faster R-CNN loss terms. The network parameters between the corresponding modules in the R-CNN and the Faster R-CNN branches are shared, including the backbone network, (modulated) deformable RoIpooling, and the 2 {\it fc} heads (the classification heads in the two branches are unshared). In inference, only the Faster R-CNN network is applied on the test images, without the auxiliary R-CNN branch. Thus, no additional computation is introduced by R-CNN feature mimicking in inference.

\section{Related Work}

\noindent\textbf{Deformation Modeling} is a long-standing problem in computer vision, and there has been tremendous effort in designing translation-invariant features. Prior to the deep learning era, notable works include scale-invariant feature transform (SIFT)~\cite{lowe1999object}, oriented FAST and rotated BRIEF (ORB)~\cite{rublee2011orb}, and deformable part-based models (DPM)~\cite{pedro2010dpm}. Such works are limited by the inferior representation power of handcrafted features and the constrained family of geometric transformations they address (\eg, affine transformations). Spatial transformer networks (STN)~\cite{Jaderberg2015} is the first work on learning translation-invariant features for deep CNNs. It learns to apply global affine transformations to warp feature maps, but such transformations inadequately model the more complex geometric variations encountered in many vision tasks. Instead of performing global parametric transformations and feature warping, Deformable ConvNets sample feature maps in a local and dense manner, via learnable offsets in the proposed deformable convolution and deformable RoIpooling modules. Deformable ConvNets is the first work to effectively model geometric transformations in complex vision tasks (\eg, object detection and semantic segmentation) on challenging benchmarks.

Our work extends Deformable ConvNets by enhancing its modeling power and facilitating network training. This new version of Deformable ConvNets yields significant performance gains over the original model.

\vspace{0.5em}
\noindent\textbf{Relation Networks and Attention Modules} are first proposed in natural language processing~\cite{gehring2016convolutional,gehring2017convolutional,britz2017massive,vaswani2017attention} and physical system modeling~\cite{battaglia2016interaction,watters2017visual,hoshen2017vain,santoro2017simple,denil2017programmable,raposo2017discovering}. An attention / relation module effects an individual element (\eg, a word in a sentence) by aggregating features from a set of elements (\eg, all the words in the sentence), where the aggregation weights are usually defined on feature similarities among the elements. They are powerful in capturing long-range dependencies and contextual information in these tasks. Recently, the concurrent works of~\cite{hu2018relation} and~\cite{wang2017non} successfully extend relation networks and attention modules to the image domain, for modeling long-range object-object and pixel-pixel relations, respectively. In~\cite{gu2018learning}, a learnable region feature extractor is proposed, unifying the previous region feature extraction modules from the pixel-object relation perspective. A common issue with such approaches is that the aggregation weights and the aggregation operation need to be computed on the elements in a pairwise fashion, incurring heavy computation that is quadratic to the number of elements (\eg, all the pixels in an image). Our developed approach can be perceived as a special attention mechanism where only a sparse set of elements have non-zero aggregation weights (\eg, $3\times 3$ pixels from among all the image pixels). The attended elements are specified by the learnable offsets, and the aggregation weights are controlled by the modulation mechanism. The computational overhead is just linear to the number of elements, which is negligible compared to that of the entire network (See Table~\ref{table:detection_frcnn}).

\vspace{0.5em}
\noindent\textbf{Spatial Support Manipulation.} For atrous convolution, the spatial support of convolutional layers has been enlarged by padding zeros in the convolutional kernels~\cite{chen2016deeplab}. The padding parameters are handpicked and predetermined. In active convolution~\cite{jeon2017active}, which is contemporary with Deformable ConvNets, convolutional kernel offsets are learned via back-propagation. But the offsets are static model parameters fixed after training and shared over different spatial locations. In a multi-path network for object detection~\cite{zagoruyko2016multipath}, multiple RoIpooling layers are employed for each input RoI to better exploit multi-scale and context information. The multiple RoIpooling layers are centered at the input RoI, and are of different spatial scales. A common issue with these approaches is that the spatial support is controlled by static parameters and does not adapt to image content.

\vspace{0.5em}
\noindent\textbf{Effective Receptive Field and Salient Region.} Towards better interpreting how a deep network functions, significant progress has been made in understanding which image regions contribute most to network prediction. Recent works on effective receptive fields~\cite{luo2017understanding} and salient regions~\cite{zhou2016learning,zintgraf2017visualizing,fong2017interpretable,dabkowski2017real} reveal that only a small proportion of pixels in the theoretical receptive field contribute significantly to the final network prediction. The effective support region is controlled by the joint effect of network weights and sampling locations. Here we exploit the developed techniques to better understand the network behavior of Deformable ConvNets. The resulting observations guide and motivate us to improve over the original model.

\vspace{0.5em}
\noindent\textbf{Network Mimicking and Distillation} are recently introduced techniques for model acceleration and compression. Given a large teacher model, a compact student model is trained by mimicking the teacher model output or feature responses on training images~\cite{ba2014deep,hinton2015distilling,li2017mimicking}. The hope is that the compact model can be better trained by distilling knowledge from the large model. 

Here we employ a feature mimic loss to help the network learn features that reflect the object focus and classification power of R-CNN features. Improved accuracy is obtained and the visualized spatial supports corroborate this approach.

\setlength{\tabcolsep}{3pt}
\renewcommand{\arraystretch}{1.0}
\begin{table*}[t]
        \centering
        \small
        \resizebox{0.95\linewidth}{!}{
        \begin{tabular}{c|l|c|ccc|cc||cc|cc}
        \Xhline{2\arrayrulewidth}
        \multirow{2}{*}{method} & \multirow{2}{*}{setting (shorter side 1000)} & \multicolumn{6}{c||}{Faster R-CNN} & \multicolumn{4}{c}{Mask R-CNN}\\
        \cline{3-12}
                & & AP$^\text{bbox}$ & AP$^\text{bbox}_\text{S}$ & AP$^\text{bbox}_\text{M}$ & AP$^\text{bbox}_\text{L}$ & param & FLOP & AP$^\text{bbox}$ & AP$^\text{mask}$ & param & FLOP \\
                \hline
                \multirow{3}{*}{baseline} & regular (RoIpooling) & 32.1 & 14.9 & 37.5 & 44.4 & 51.3M & 326.7G & - & -  & - & -\\
                & regular (aligned RoIpooling) & 34.7 & 19.3 & 39.5 & 45.3 & 51.3M & 326.7G & 36.6 & 32.2 & 39.5M & 447.5G\\
                & dconv@c5 + dpool (DCNv1) & 38.0 & 20.7 & 41.8 & 52.2 & 52.7M & 328.2G & 40.4 & 35.3 & 40.9M & 449.0G \\
                \hline
                \multirow{5}{*}{\makecell{enriched \\deformation}}& dconv@c5 & 37.4 & 20.0 & 40.9 & 51.0 & 51.5M & 327.1G & 40.2 & 35.1 & 39.8M & 447.8G \\
                & dconv@c4$\sim$c5 & 40.0 & 21.4 & 43.8 & 55.3 & 51.7M & 328.6G & 41.8 & 36.8 & 40.0M & 449.4G \\
                & dconv@c3$\sim$c5 & 40.4 & 21.6 & 44.2 & 56.2 & 51.8M & 330.6G & 42.2 & 37.0 & 40.1M & 451.4G \\
                & dconv@c3$\sim$c5 + dpool & 41.0 & 22.0 & 45.1 & 56.6 & 53.0M & 331.8G & 42.4 & 37.0 & 41.3M & 452.5G \\
                & mdconv@c3$\sim$c5 + mdpool & 41.7 & 22.2 & 45.8 & 58.7 & 65.5M & 346.2G & 43.1 & 37.3 & 53.8M & 461.1G \\
                \Xhline{2\arrayrulewidth}
        \end{tabular}
        }
        \vspace{.5em}
        \caption{Ablation study on enriched deformation modeling. The input images are of shorter side 1,000 pixels (default in paper). In the setting column, ``(m)dconv'' and ``(m)dpool'' stand for (modulated) deformable convolution and (modulated) deformable RoIpooling, respectively. Also, ``dconv@c3$\sim$c5'' stands for applying deformable conv layers at stages conv3$\sim$conv5, for example. Results are reported on the COCO\ 2017 validation set.}
        \label{table:detection_frcnn}
\end{table*}

\setlength{\tabcolsep}{3pt}
\renewcommand{\arraystretch}{1.0}
\begin{table*}[t]
        \centering
        \small
        \resizebox{0.95\linewidth}{!}{
        \begin{tabular}{c|l|c|ccc|cc||cc|cc}
        \Xhline{2\arrayrulewidth}
        \multirow{2}{*}{method} & \multirow{2}{*}{setting (shorter side 800)} & \multicolumn{6}{c||}{Faster R-CNN} & \multicolumn{4}{c}{Mask R-CNN}\\
        \cline{3-12}
                & & AP$^\text{bbox}$ & AP$^\text{bbox}_\text{S}$ & AP$^\text{bbox}_\text{M}$ & AP$^\text{bbox}_\text{L}$ & param & FLOP & AP$^\text{bbox}$ & AP$^\text{mask}$ & param & FLOP \\
                \hline
                \multirow{3}{*}{baseline} & regular (RoIpooling) & 32.8 & 13.6 & 37.2 & 48.7 & 51.3M & 196.8G & - & -  & - & -\\
                & regular (aligned RoIpooling) & 35.6 & 18.2 & 40.3 & 48.7 & 51.3M & 196.8G & 37.8 & 33.4  & 39.5M & 303.5G\\
                & dconv@c5 + dpool (DCNv1) & 38.2 & 19.1 & 42.2 & 54.0 & 52.7M & 198.9G & 40.3 & 35.0 & 40.9M & 304.9G \\
                \hline
                \multirow{5}{*}{\makecell{enriched \\deformation}}& dconv@c5 & 37.6 & 19.3 & 41.4 & 52.6 & 51.5M & 197.7G & 39.9 & 34.9 & 39.8M & 303.7G \\
                & dconv@c4$\sim$c5 & 39.2 & 19.9 & 43.4 & 55.5 & 51.7M & 198.7G & 41.2 & 36.1 & 40.0M & 304.7G \\
                & dconv@c3$\sim$c5 & 39.5 & 21.0 & 43.5 & 55.6 & 51.8M & 200.0G & 41.5 & 36.4 & 40.1M & 306.0G \\
                & dconv@c3$\sim$c5 + dpool & 40.0 & 21.1 & 44.6 & 56.3 & 53.0M & 201.2G & 41.8 & 36.4 & 41.3M & 307.2G \\
                & mdconv@c3$\sim$c5 + mdpool & 40.8 & 21.3 & 45.0 & 58.5 & 65.5M & 214.7G & 42.7 & 37.0 & 53.8M & 320.3G \\
                \Xhline{2\arrayrulewidth}
        \end{tabular}
        }
        \vspace{.5em}
        \caption{Ablation study on enriched deformation modeling. The input images are of shorter side 800 pixels. Results are reported on the COCO\ 2017 validation set.}
        \label{table:detection_frcnn_800}
\end{table*}

\section{Experiments}

\subsection{Experiment Settings}
\label{sec:exp_setting}

Our models are trained on the 118k images of the COCO 2017 train set. In ablation, evaluation is done on the 5k images of the COCO 2017 validation set. We also evaluate performance on the 20k images of the COCO 2017 test-dev set. The standard mean average-precision scores at different box and mask IoUs are used for measuring object detection and instance segmentation accuracy, respectively.

Faster R-CNN and Mask R-CNN are chosen as the baseline systems. ImageNet~\cite{deng2009imagenet} pre-trained ResNet-50 is utilized as the backbone. The implementation of Faster R-CNN is the same as in Section~\ref{sec:mimic}. For Mask R-CNN, we follow the implementation in~\cite{he2017mask}. To turn the networks into their deformable counterparts, the last  set of $3\times 3$ regular conv layers (close to the output in the bottom-up computation) are replaced by (modulated) deformable conv layers. Aligned RoIpooling is replaced by (modulated) deformable RoIpooling. Specially for Mask R-CNN, the two aligned RoIpooling layers with $7\times 7$ and $14\times 14$ bins are replaced by two (modulated) deformable RoIpooling layers with the same bin numbers. In R-CNN feature mimicking, the feature mimic loss is enforced on the RoI head for classification only (excluding that for mask estimation). For both systems, the choice of hyper-parameters follows the latest Detectron~\cite{Detectron2018} code base except for the image resolution, which is briefly presented here. In both training and inference, images are resized so that the shorter side is 1,000 pixels\footnote{The previous default setting in Detectron is 800 pixels. Ablation on input image resolution is present in Appendix.}. Anchors of 5 scales and 3 aspect ratios are utilized. 2k and 1k region proposals are generated at a non-maximum suppression threshold of 0.7 at training and inference respectively. In SGD training, 256 anchor boxes (of positive-negative ratio 1:1) and 512 region proposals (of positive-negative ratio 1:3) are sampled for backpropagating their gradients. In our experiments, the networks are trained on 8 GPUs with 2 images per GPU for 16 epochs. The learning rate is initialized to 0.02 and is divided by 10 at the 10-th and the 14-th epochs. The weight decay and the momentum parameters are set to $10^{-4}$ and 0.9, respectively.

\subsection{Enriched Deformation Modeling}
\label{sec:exp_enriched}

The effects of enriched deformation modeling are examined from ablations shown in Table~\ref{table:detection_frcnn}. The baseline with regular CNN modules obtains an AP$^{\text{bbox}}$ score of 34.7\% for Faster R-CNN, and AP$^{\text{bbox}}$ and AP$^{\text{mask}}$ scores of 36.6\% and 32.2\% respectively for Mask R-CNN. To obtain a DCNv1 baseline, we follow the original Deformable ConvNets paper by replacing the last three layers of $3\times 3$ convolution in the conv5 stage and the aligned RoIpooling layer by their deformable counterparts. This DCNv1 baseline achieves an AP$^{\text{bbox}}$ score of 38.0\% for Faster R-CNN, and AP$^{\text{bbox}}$ and AP$^{\text{mask}}$ scores of 40.4\% and 35.3\% respectively for Mask R-CNN. The deformable modules considerably improve accuracy as observed in~\cite{dai2017deformable}.

By replacing more $3\times 3$ regular conv layers by their deformable counterparts, the accuracy of both Faster R-CNN and Mask R-CNN steadily improve, with gains between 2.0\% and 3.0\% for AP$^{\text{bbox}}$ and AP$^{\text{mask}}$ scores when the conv layers in conv3-conv5 are replaced. No additional improvement is observed on the COCO benchmark by further replacing the regular conv layers in the conv2 stage. By upgrading the deformable modules to modulated deformable modules, we obtain further gains between 0.3\% and 0.7\% in AP$^{\text{bbox}}$ and AP$^{\text{mask}}$ scores. In total, enriching the deformation modeling capability yields a 41.7\% AP$^{\text{bbox}}$ score on Faster R-CNN, which is 3.7\% higher than that of the DCNv1 baseline. On Mask R-CNN, 43.1\% AP$^{\text{bbox}}$ and 37.3\% AP$^{\text{mask}}$ scores are obtained with the enriched deformation modeling, which are respectively 2.7\% and 2.0\% higher than those of the DCNv1 baseline. Note that the added parameters and FLOPs for enriching the deformation modeling are minor compared to those of the overall networks.

Shown in Figure~\ref{fig:visualize_regular_deformable_conv} (b)$\sim$(c), the spatial support of the enriched deformable modeling exhibits better adaptation to image content compared to that of DCNv1. 

Table~\ref{table:detection_frcnn_800} presents the results at input image resolution of 800 pixels, which follows the default setting in the Detectron code base. The same conclusion holds.

\subsection{R-CNN Feature Mimicking}
\label{sec:exp_mimicking}

Ablations of the design choices in R-CNN feature mimicking are shown in Table~\ref{table:detection_modulation_mimicking}. With the enriched deformation modeling, R-CNN feature mimicking further improves the AP$^{\text{bbox}}$ and AP$^{\text{mask}}$ scores by about 1\% to 1.4\% in both the Faster R-CNN and Mask R-CNN systems. Mimicking features of positive boxes on the object foreground is found to be particularly effective, and the results when mimicking all the boxes or just negative boxes are much lower. As shown in Figure~\ref{fig:visualize_regular_deformable_roi} (c)$\sim$(d), feature mimicking can help the network features better focus on the object foreground, which is beneficial for positive boxes. For the negative boxes, the network tends to exploit more context information (see Figure~\ref{fig:visualize_regular_deformable_roi}), where feature mimicking would not be helpful.

We also apply R-CNN feature mimicking to regular ConvNets without any deformable layers. Almost no accuracy gains are observed. The visualized spatial support regions are shown in Figure~\ref{fig:visualize_regular_deformable_roi} (e), which are not focused on the object foreground even with the auxiliary mimic loss. This is likely because it is beyond the representation capability of regular ConvNets to focus features on the object foreground, and thus this cannot be learned.

\setlength{\tabcolsep}{3pt}
\renewcommand{\arraystretch}{1}
\begin{table}[t]
        \centering
        \small
        \resizebox{\linewidth}{!}{
        \begin{tabular}{c|c|c|cc}
        \Xhline{2\arrayrulewidth}
        \multirow{2}{*}{\makecell{setting}} & \multirow{2}{*}{\makecell{regions to\\mimic}} & \multicolumn{1}{c|}{\makecell{Faster\\R-CNN}} & \multicolumn{2}{c}{\makecell{Mask\\R-CNN}} \\
        \cline{3-5}
                 & & AP$^\text{bbox}$ & AP$^\text{bbox}$ & AP$^\text{mask}$ \\
                 \hline
                 \multirow{4}{*}{\makecell{mdconv3$\sim$5 + \\ mdpool}} & None & 41.7 & 43.1 & 37.3  \\
                 & FG \& BG & 42.1 & 43.4 & 37.6 \\
                 & BG Only & 41.7 & 43.3 & 37.5 \\
                 & FG Only & 43.1 & 44.3 & 38.3 \\
                \hline
         \multirow{2}{*}{regular} & None & 34.7 & 36.6 & 32.2 \\
                 & FG Only & 35.0 & 36.8 & 32.3 \\
              \Xhline{2\arrayrulewidth}
        \end{tabular}}
        \vspace{.5em}
        \caption{Ablation study on R-CNN feature mimicking. Results are reported on the COCO\ 2017 validation set.}
        \label{table:detection_modulation_mimicking}
\end{table}

\setlength{\tabcolsep}{4pt}
\renewcommand{\arraystretch}{1}
\begin{table}[t]
        \centering
        \small
        \resizebox{0.95\linewidth}{!}{
        \begin{tabular}{c|c|c|cc}
        \Xhline{2\arrayrulewidth}
         \multirow{2}{*}{\makecell{backbone}} & \multirow{2}{*}{method} & \multicolumn{1}{c|}{\makecell{Faster\\R-CNN}} & \multicolumn{2}{c}{\makecell{Mask\\R-CNN}} \\
                 \cline{3-5}
                 & & AP$^\text{bbox}$ & AP$^\text{bbox}$ & AP$^\text{mask}$ \\
                 \hline             \multirow{3}{*}{\makecell{ResNet-50}} & regular & 35.1 & 37.0 & 32.4 \\
                 & DCNv1 & 38.4 & 40.7 & 35.5 \\
                 & DCNv2 & 43.3 & 44.5 & 38.4 \\
        \hline
        \multirow{3}{*}{\makecell{ResNet-101}} & regular & 39.2 & 40.9 & 35.3 \\
                 & DCNv1 & 41.4 & 42.9 & 37.1 \\
                 & DCNv2 & 44.8 & 45.8 & 39.7 \\
                 \hline            \multirow{3}{*}{\makecell{ResNext-101}} & regular & 40.1 & 41.7 & 36.2 \\
                 & DCNv1 & 41.7 & 43.4 & 37.7 \\
                 & DCNv2 & 45.3 & 46.7 & 40.5 \\
                \Xhline{2\arrayrulewidth}
        \end{tabular}}
        \vspace{.5em}
        \caption{Results of DCNv2, DCNv1 and regular ConvNets on various backbones on the COCO 2017 test-dev set.}
        \label{table:strong_backbone}
        \vspace{-1em}
\end{table}

\subsection{Application on Stronger Backbones}

Results on stronger backbones, by replacing ResNet-50 with ResNet-101 and ResNext-101~\cite{xie2017aggregated}, are presented in Table~\ref{table:strong_backbone}. For the entries of DCNv1, the regular $3 \times 3$ conv layers in the conv5 stage are replaced by the deformable counterpart, and aligned RoIpooling is replaced by deformable RoIpooling. For the DCNv2 entries, all the $3 \times 3$ conv layers in the conv3-conv5 stages are of modulated deformable convolution, and modulated deformable RoIpooling is used instead, with supervision from the R-CNN feature mimic loss. DCNv2 is found to outperform regular ConvNet and DCNv1 considerably on all the network backbones.

\section{Conclusion}

Despite the superior performance of Deformable ConvNets in modeling geometric variations, its spatial support extends well beyond the region of interest, causing features to be influenced by irrelevant image content. In this paper, we present a reformulation of Deformable ConvNets which improves its ability to focus on pertinent image regions, through increased modeling power and stronger training. Significant performance gains are obtained on the COCO benchmark for object detection and instance segmentation.

\appendix
\renewcommand{\thesection}{A\arabic{section}}  

\begin{figure*}[t] 
  \centering 
  \subfigure[AP$^\text{bbox}$ for all objects]{ 
    \includegraphics[width=0.49\linewidth]{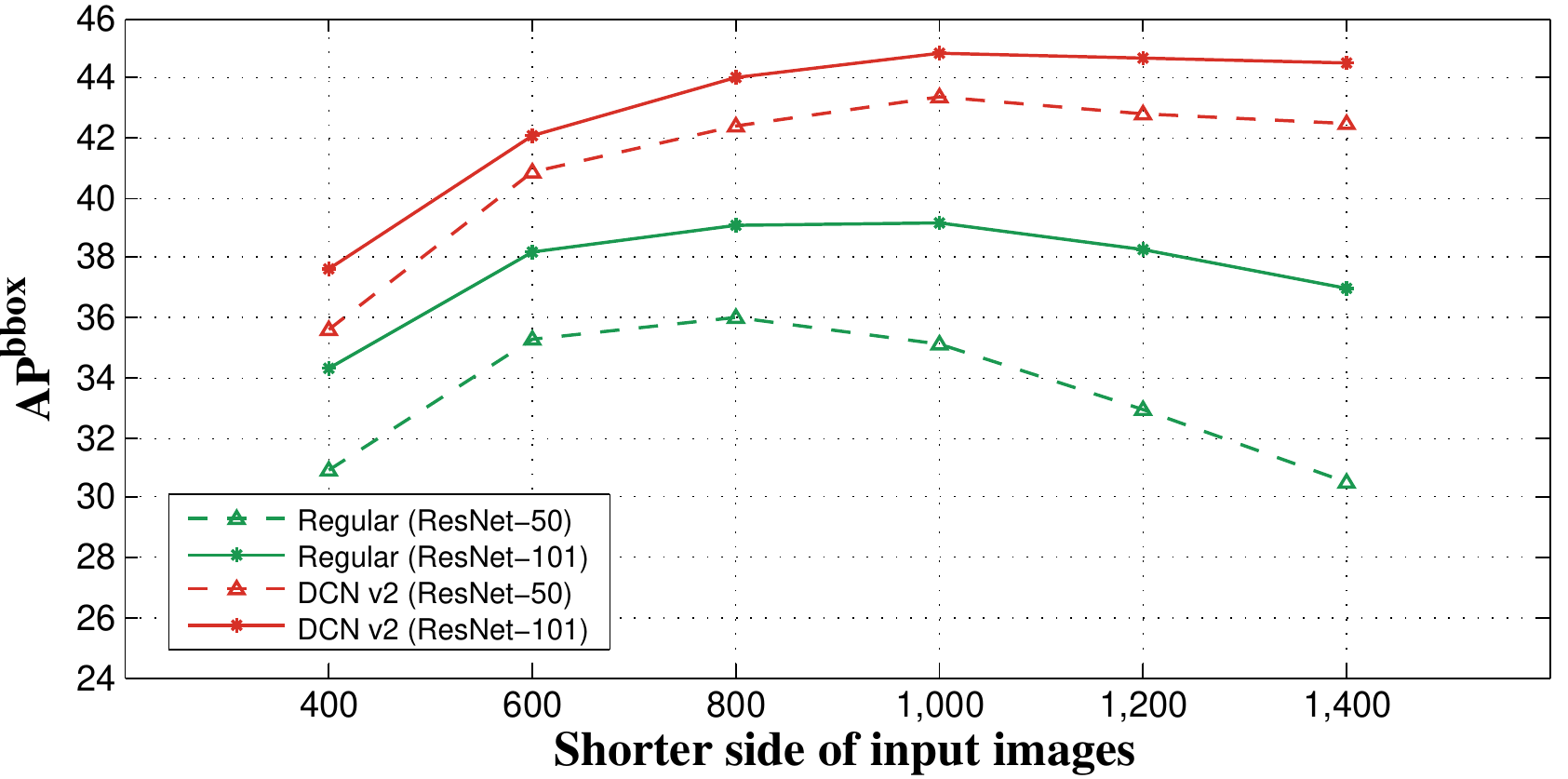}}  
  \subfigure[AP$^\text{bbox}_{\text{S}}$ for small objects]{
    \includegraphics[width=0.49\linewidth]{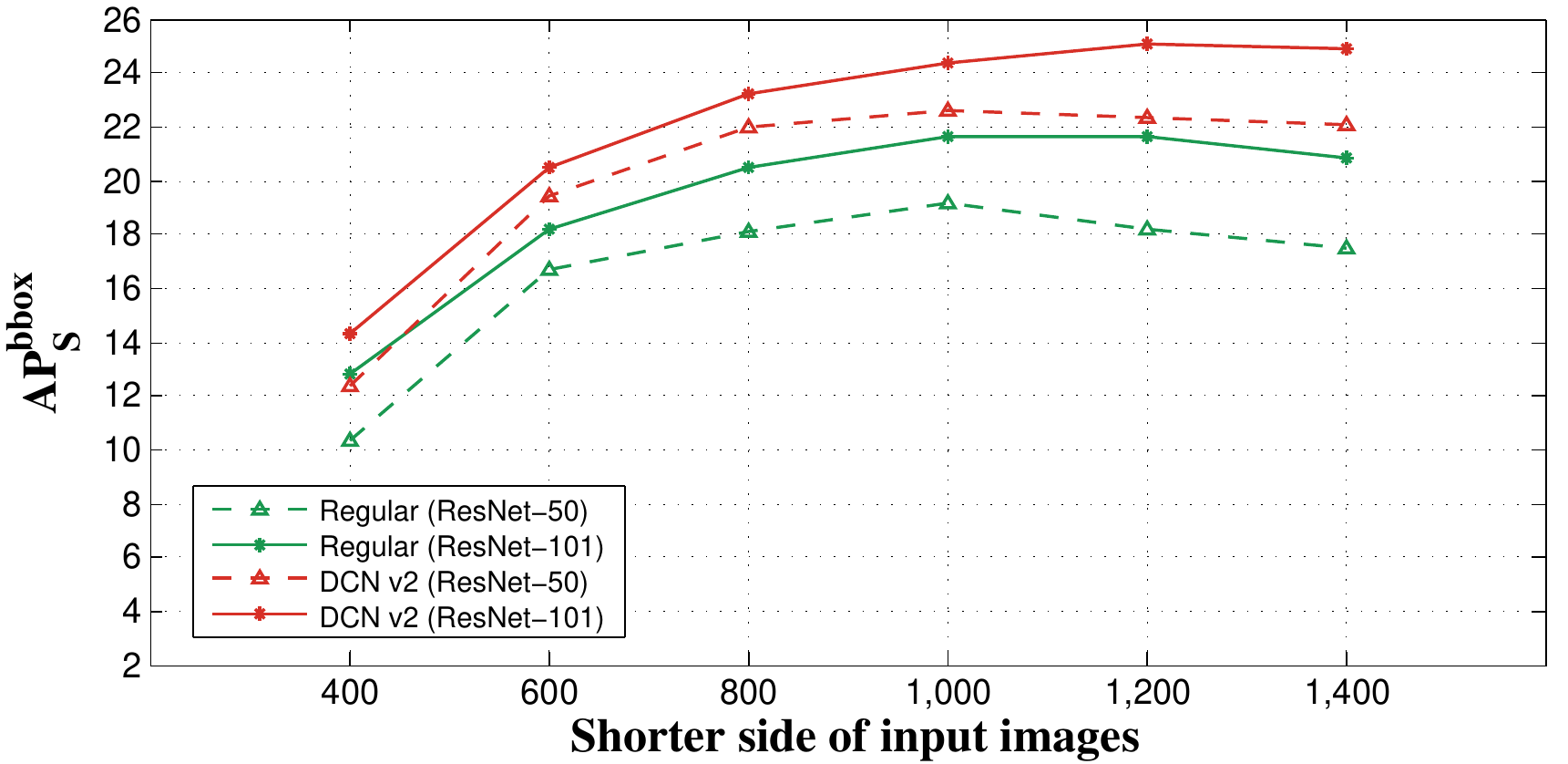}}
  \subfigure[AP$^\text{bbox}_{\text{M}}$ for medium objects]{
    \includegraphics[width=0.49\linewidth]{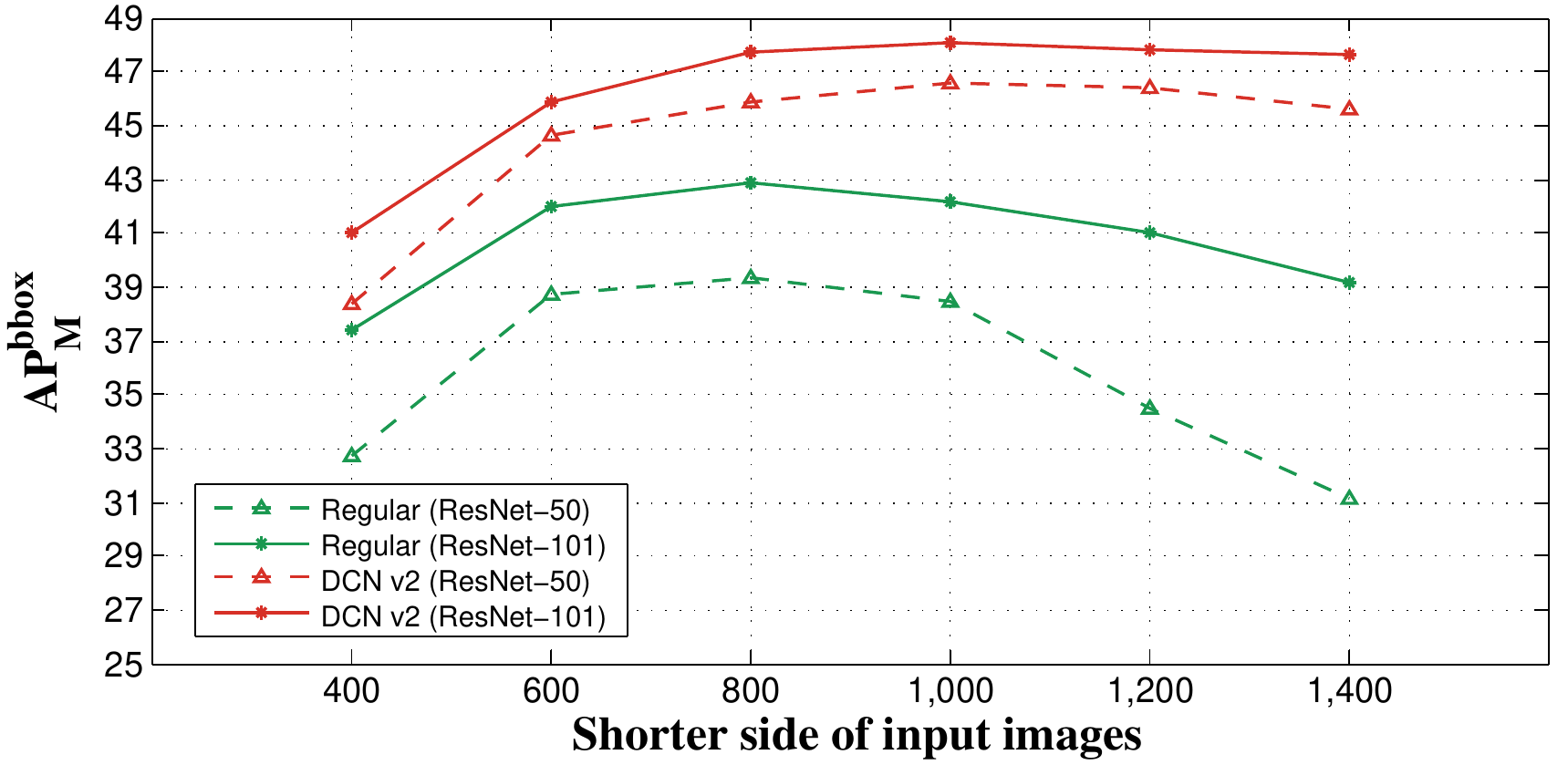}}
  \subfigure[AP$^\text{bbox}_{\text{L}}$ for large objects]{
    \includegraphics[width=0.49\linewidth]{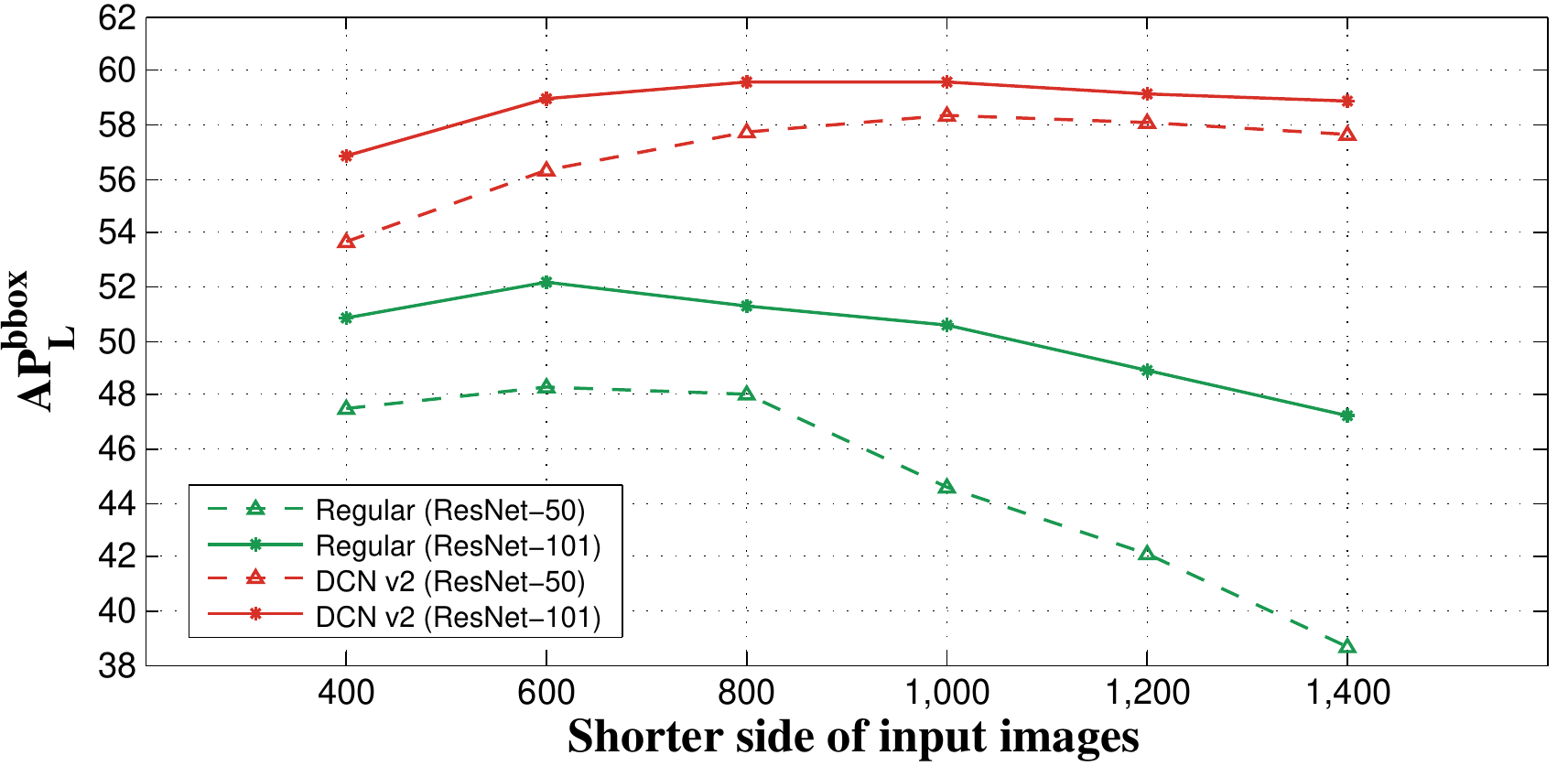}}

  \caption{AP$^\text{bbox}$ scores of DCNv2 and regular ConvNets (Faster R-CNN + ResNet-50 / ResNet-101) on  input images of varies resolution on the COCO 2017 test-dev set.}
  \label{fig:multi_scale_model}

  \vspace{-1em}
\end{figure*}

\begin{figure*}[t]
        \centering
        \subfigure[regular conv]{ 
          \includegraphics[width=0.49\linewidth]{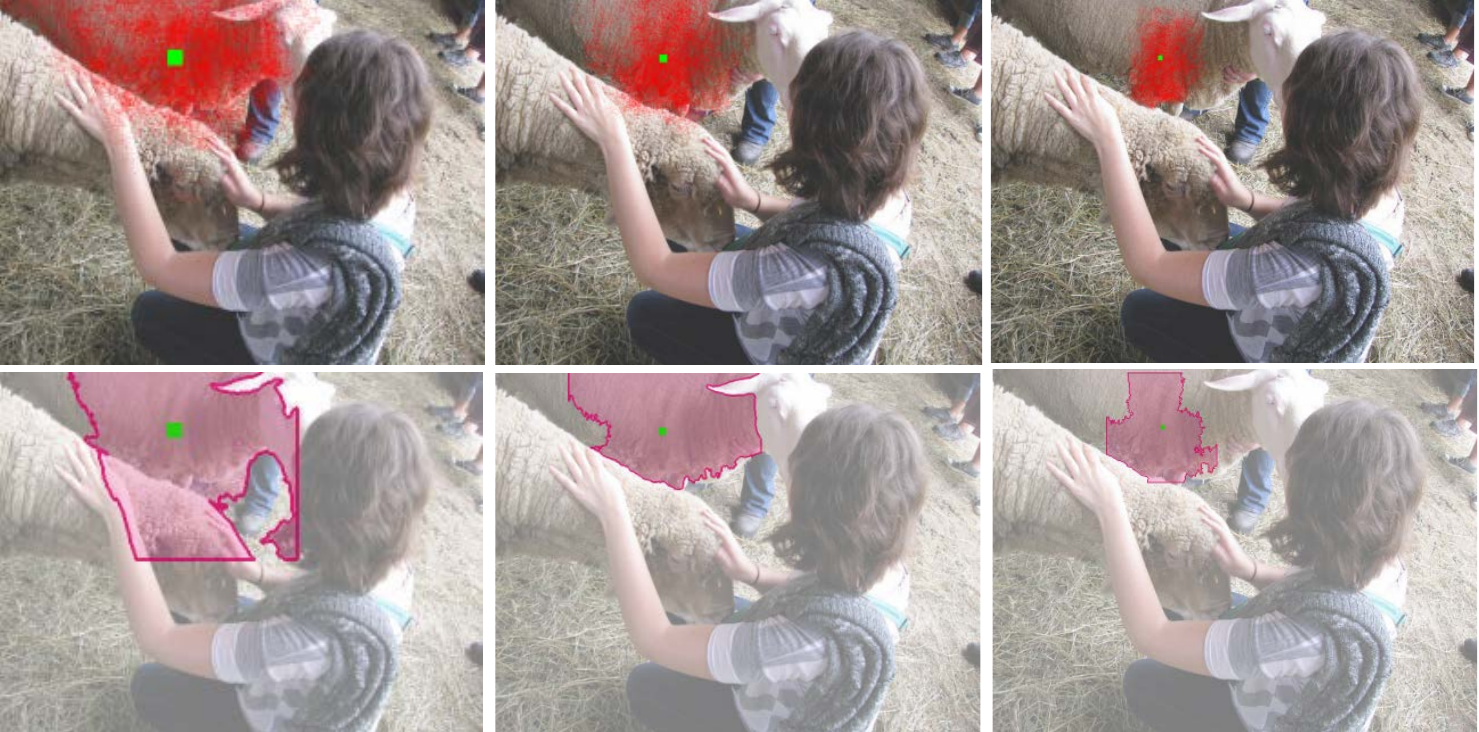}}  
        \subfigure[modulated deformable conv@conv3$\sim$5 stages (DCNv2)]{
          \includegraphics[width=0.49\linewidth]{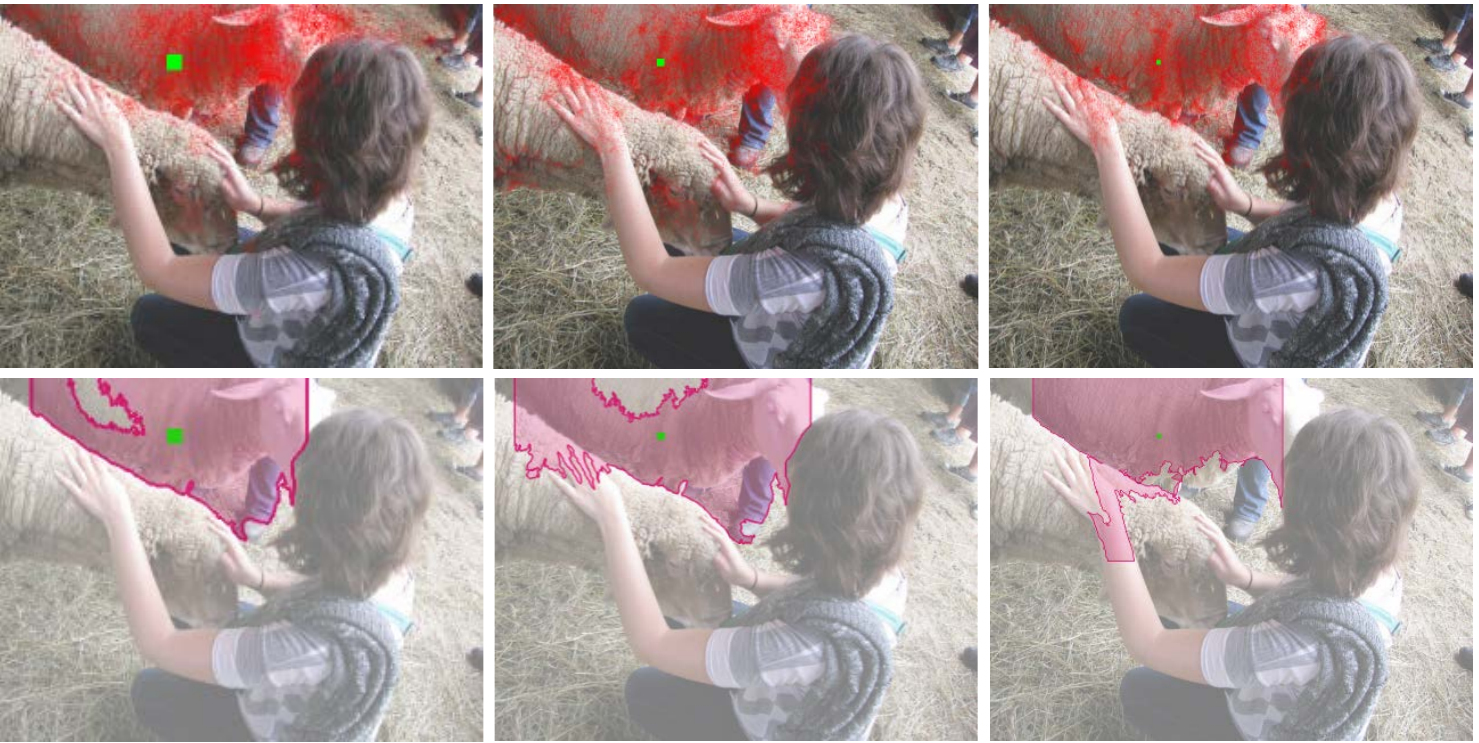}}
        \caption{Spatial support of nodes in the last layer of the conv5 stage in DCNv2 and regular ConvNets. Input images are of shorter side 400 pixels (left), 800 pixels (middle), and 1400 pixels (right), respectively. The effective receptive field and error-bounded saliency regions are shown in the top and bottom rows, respectively.}
        \label{fig:visualize_image_scale}
\end{figure*}

\setlength{\tabcolsep}{3pt}
\renewcommand{\arraystretch}{1.2}
\begin{table*}[t]
        \centering
        \small
        \begin{tabular}{c|l|ccc|ccc}
        \Xhline{2\arrayrulewidth}
        \multirow{2}{*}{method} & \multirow{2}{*}{setting} & \multicolumn{6}{c}{Faster R-CNN + ResNet-101}\\
        \cline{3-8}
                & & AP$^\text{bbox}$ & AP$^\text{bbox}_\text{50}$ & AP$^\text{bbox}_\text{75}$ & AP$^\text{bbox}_\text{S}$ & AP$^\text{bbox}_\text{M}$ & AP$^\text{bbox}_\text{L}$\\
                \hline
                \multirow{3}{*}{regular} & single-scale, shorter side 800 & 39.1 & 60.6 & 42.2 & 20.5 & 42.9 & 51.3 \\
                & single-scale, shorter side 1000 (best) & 39.2 & 60.6 & 42.4 & 21.6 & 42.2 & 51.3 \\
                \cline{2-8}
                & multi-scale test & 41.2 & 62.4 & 45.2 & 24.6 & 44.3 & 52.7 \\
                \hline
                \multirow{3}{*}{DCNv2}& single-scale, shorter side 800 & 44.0 & 65.9 & 48.1 & 23.2 & 47.7 & 59.6 \\
                & single-scale, shorter side 1000 (best) & 44.8 & 66.3 & 48.8 & 24.4 & 48.1 & 59.6 \\
                \cline{2-8}
                & multi-scale test & 46.0 & 67.9 & 50.8 & 27.8 & 49.1 & 59.5 \\
                \Xhline{2\arrayrulewidth}
        \end{tabular}
        \vspace{.5em}
        \caption{Ablation study on input image resolution. Results are reported on the COCO 2017 test-dev set.}
        \label{table:detection_multi_scale}
        \vspace{-1em}
\end{table*}

\begin{table}
        \centering
        \begin{tabular}{c|c|cc|cc}
        \Xhline{2\arrayrulewidth}
                  backbone & method & \makecell{top-1\\acc (\%)} &  \makecell{top-5\\acc (\%)} & param & FLOP\\
                 \hline
                 \multirow{3}{*}{ResNet-50} & regular & 76.5 & 93.1 & 26.6M & 4.1G\\
                 & DCNv1 & 76.6 & 93.2 & 26.8M & 4.1G\\
                 & DCNv2 & 78.2 & 94.0 & 27.4M & 4.3G\\
                 \hline
                 \multirow{3}{*}{ResNet-101} & regular & 78.4 & 94.2 & 45.5M & 7.8G \\
                 & DCNv1 & 78.4 & 94.2 & 45.8M & 7.8G \\
                 & DCNv2 & 79.2 & 94.6 & 47.4M & 8.2G \\
                 \hline
                 \multirow{3}{*}{ResNeXt-101} & regular & 78.8 & 94.4 & 45.1M & 8.0G\\
                 & DCNv1 & 78.9 & 94.4 & 45.6M & 8.0G \\
                 & DCNv2 & 79.8 & 94.8 & 49.0M & 8.7G \\
                \Xhline{2\arrayrulewidth}
        \end{tabular}
        \vspace{.5em}
        \caption{ImageNet classification accuracies of DCNv2, DCNv1 and regular ConvNets.}
        \label{table:imagenet}
\end{table}

\setlength{\tabcolsep}{4pt}
\renewcommand{\arraystretch}{1.2}
\begin{table*}
        \centering
        \small
        \begin{tabular}{c|c|cc|c|c|c}
        \Xhline{2\arrayrulewidth}
                 \multirow{2}{*}{method} & \multirow{2}{*}{\makecell{offset\&modulation\\pretraining}} & \multicolumn{2}{c|}{VOC det} & \multicolumn{1}{c|}{VOC seg} & \multicolumn{1}{c|}{ImageNet VID det} & \multicolumn{1}{c}{COCO det} \\
                 \cline{3-7}
                 & & AP$_{50}^\text{bbox}$ & AP$_{70}^\text{bbox}$ & mIoU & AP$^\text{bbox}$ & AP$^\text{bbox}$ \\
                 \hline
                 regular & none & 81.9 & 68.2 & 72.0 & 74.9 & 39.2\\
                 DCNv2 & none & 83.7 & 72.4 & 76.1 & 79.2 & 44.8\\
                 DCNv2 & ImageNet & 84.9 & 73.5 & 78.3 & 80.7 & 44.9\\
                \Xhline{2\arrayrulewidth}
        \end{tabular}
        \vspace{.5em}
        \caption{Finetuning the ImageNet-pretrained DCNv2 model for various tasks and benchmarks. ResNet-101 is utilized as the backbone.}
        \label{table:finetune}
\end{table*}

\section{Error-bounded Image Saliency}

In existing research on image saliency~\cite{zhou2016learning,zintgraf2017visualizing,fong2017interpretable,dabkowski2017real}, a widely utilized formulation is as follows. Given an input image $\mathbf{I}$ and a trained network $\mathcal{N}$, let $\mathcal{N}(\mathbf{I})$ denote the network response on the original image. A binary mask $M$, which is of the same spatial dimension as $\mathbf{I}$, can be applied on the image as $\mathbf{I} \odot M$. For the image pixel $p$ where $M(p)=1$, its content is kept in the masked image. Meanwhile, if $M(p)=0$, the content is set as 0 in the masked image. The saliency map is obtained by optimizing loss function $L(M) = ||\mathcal{N}(\mathbf{I})-\mathcal{N}(\mathbf{I}\odot M)||_2 + \lambda ||M||_1$ as a function of $M$, where $\lambda$ is the hyper-parameter balancing the output reconstruction error $||\mathcal{N}(\mathbf{I})-\mathcal{N}(\mathbf{I}\odot M)||_2$ and the salient area loss $||M||_1$. The optimized mask $M$ is called the saliency map. The problem is it is hard to obtain the salient region at a specified reconstruction error. Thus it is hard to compare the salient regions from two networks at the same reconstruction error.

We seek to strictly constrain the reconstruction loss in the image saliency formulation, so as to facilitate comparison among the salient regions derived from different networks. Thus, the optimization problem is slightly modified to be
\begin{equation}
\begin{aligned}
\min & ||M||_1 \\
\text{s.t.} \quad L_{\text{rec}}(\mathcal{N}(\mathbf{I}), \mathcal{N}(& \mathbf{I}\odot M)) < \epsilon,
\end{aligned}
\label{eq:visual_support_formulation}
\end{equation}
where $L_{\text{rec}}(\mathcal{N}(\mathbf{I}), \mathcal{N}(\mathbf{I}\odot M))$ denotes an arbitrary form of reconstruction loss, which is strictly bounded by $\epsilon$. We term the collection of image pixels where $\{p|M(p)=1\}$ in the optimized mask as visual support region.

The formulation in Eq.~\eqref{eq:visual_support_formulation} is hard to be optimized, due to the hard reconstruction error constraint introduced. Here we develop a heuristic two-step procedure to reduce the search space in deriving the visual support region. At the first step, the visual support region is constrained to be rectangular of arbitrary shape. The rectangular is centered on the node to be interpreted. The rectangular is initialized of area size 0, and is enlarged gradually (at even area increment). The enlargement stops upon the reconstruction error constraint is satisfied. At the second step, pixel-level visual support region is derived within the rectangular area. The image is segmented into super-pixels by the algorithm in \cite{achanta2012slic}, so as to restrict the solution space. At initial, all the super-pixels within the rectangular are counted in the visual support region (taking mask value 1). Then the super-pixels are gradually removed in a greedy manner. At each iteration, the super-pixel causing the smallest rise in reconstruction error is removed. The iteration stops till the constraint would be violated by removing anymore super-pixels.

We apply the two-step procedure to visualize network nodes in Faster R-CNN object detector~\cite{ren2015faster}. We visualize both feature map nodes shared on the whole image, and the 2{\it fc} node in the per-RoI detection head, which is directly followed by the classification and the bounding box regression branches. For image-wise feature map nodes (at a certain location), square rectangular is applied in the two-step procedure. For RoI-wise feature nodes, the rectangular is of the same aspect ratio as the input RoI. 
For both image-wise and RoI-wise nodes, the reconstruction loss is one minus the cosine similarity between the feature vectors derived from masked and original images. The error upper bound $\epsilon$ is set as 0.1. 

\section{DCNv2 with Various Image Resolution}

Figure~\ref{fig:multi_scale_model} presents the results of applying regular ConvNets and DCNv2 on images of various resolutions. The baseline model is Faster R-CNN with ResNet-50~\cite{he2016deep} and ResNet-101. Models are trained and applied on images of shorter side $\{400, 600, 800, 1000, 1200, 1400\}$ pixels, respectively. DCNv2 is found to outperform regular ConvNet on all input resolutions. For DCNv2, the highest AP$^\text{bbox}$ scores are obtained at input images of shorter side 1,000 pixels. With the shorter side larger than 1,000 pixels, AP$^\text{bbox}$ scores of regular ConvNet decrease noticeably, while those of DCNv2 are almost unchanged. This phenomenon is more obvious for objects of large and medium sizes. As shown in Figure~\ref{fig:visualize_image_scale}, the spatial support of regular ConvNets can just cover a small portion of the large objects at such high resolution, and the accuracy suffers. Meanwhile, the spatial support of DCNv2 can effectively adapt to objects at various resolutions.

Table~\ref{table:detection_multi_scale} presents the results of multi-scale testing using ResNet-101. We first apply the DCNv2 model trained on the best single-scale setting (shorter side of 1000 pixels) on multi-scale testing images. Following the latest Detectron~\cite{Detectron2018} code base, test images range from shorter side of 400 to 1400 pixels with step size of 200 pixels. Multi-scale testing of DCNv2 improves the AP$^\text{bbox}$ score by 1.2\% compared with the best single-scale setting.

\section{ImageNet Pre-Trained DCNv2}

It is well known that many vision tasks benefit from ImageNet pre-training. This section investigates pre-training the learnable offsets and modulation scalars of DCNv2 on ImageNet~\cite{deng2009imagenet}, and finetuning on several tasks. 

\paragraph{ImageNet Pretraining} DCNv2 together with its DCNv1~\cite{dai2017deformable} and regular ConvNet counterparts are pretrained on the ImageNet-1K training set. In experiments, we follow~\cite{xie2017aggregated} for the training and inference settings. In DCNv1, the $3\times 3$ conv layers in the conv5 stage are replaced by deformable conv layers. In DCNv2, all the $3\times 3$ conv layers in the conv3$\sim$conv5 stages are replaced by modulated deformable conv layers.

Table~\ref{table:imagenet} presents the top-1 and top-5 classification accuracies on the validation set. DCNv2 achieves noticeable improvements over both the regular and DCNv1 baselines, with minor additional computation overhead. The enriched deformation modeling capability of DCNv2 is beneficial for the ImageNet classification task itself.

\paragraph{Fine-tuning for Specific Tasks} We investigate the effect of ImageNet pretrained DCNv2 models on several tasks, including object detection on Pascal VOC, ImageNet VID and COCO, and semantic segmentation on Pascal VOC\footnote{Note the mimicking module is not involved in semantic segmentation experiments.}. In experiments, Faster R-CNN and DeepLab-v1~\cite{chen2016deeplab} are adopted as the baseline systems for object detection and semantic segmentation, respectively. For object detection on COCO, we follow the same settings as in Section~\ref{sec:exp_setting}. For experiments on Pascal VOC, we mainly follow \cite{dai2017deformable} for the training and inference settings. Note that the baseline accuracy is higher than that reported in \cite{dai2017deformable} mainly because of a better ImageNet pretrained model and the introduction of RoIAlign in object detection. For object detection on ImageNet VID, we mainly follow the protocol in~\cite{lee2016multi,zhu2017deep,zhu2017flow} for the training and inference settings. The details are presented at the end of this section.
 
Table~\ref{table:finetune} compares the performance of DCNv2 on various tasks using different pre-trained models. By pre-training the learnable offsets and modulation scalars on ImageNet, rather than initializing them as zeros prior to fine-tuning, noticeably accuracy improvements are observed on PASCAL VOC object detection and semantic segmentation. Meanwhile, the effect of pre-training on COCO detection is minor. This is probably because the larger and more challenging benchmark of COCO is sufficient for learning the offsets and the modulation scalars from scratch.

\emph{ImageNet VID settings.} The models are trained on the union of the ImageNet VID training set and the ImageNet DET training set (only the same 30 category labels are used), and are evaluated on the ImageNet VID validation set. In both training and inference, the input images are resized to a shorter side of 600 pixels. In RPN, the anchors are of 3 aspect ratios $\{$1:2, 1:1, 2:1$\}$ and 4 scales $\{64^2, 128^2, 256^2, 512^2\}$. 300 region proposals are generated for each frame at an NMS threshold of 0.7 IoU. SGD training is performed, with one image at each mini-batch. 120k iterations are performed on 4 GPUs, with each GPU holding one mini-batch. The learning rates are $10^{-3}$ and $10^{-4}$ in the ﬁrst 80k and last 40k iterations,respectively. In each mini-batch,images are sampled from ImageNet DET and ImageNet VID at a 1:1 ratio. The weight decay and the momentum parameters are set to 0.0001 and 0.9, respectively. In inference, detection boxes are generated at an NMS threshold of 0.3 IoU.

{\small
\bibliographystyle{ieee}
\bibliography{egbib}
}

\end{document}